\documentclass[preprint,12pt]{elsarticle}

\usepackage{amsmath,amssymb,amsthm}
\usepackage{mathtools}
\usepackage{algorithm}
\usepackage{algpseudocode}
\usepackage{booktabs}
\usepackage{array}
\usepackage{multirow}
\usepackage{graphicx}
\usepackage{xcolor}
\usepackage{bm}
\usepackage{lineno}
\usepackage[hidelinks]{hyperref}
\usepackage{rotating}
\usepackage{placeins}
\usepackage[T1]{fontenc}
\usepackage{lmodern}
\usepackage{adjustbox}
\usepackage{caption}
\newsavebox\FrozenTabBox
\let\FrozenTabular\tabular \let\endFrozenTabular\endtabular
\renewenvironment{tabular}{\begin{lrbox}{\FrozenTabBox}\FrozenTabular}{\endFrozenTabular\end{lrbox}%
  \ifdim\wd\FrozenTabBox>\linewidth\typeout{[tabscale] Table \thetable: \the\wd\FrozenTabBox\space > \the\linewidth}\fi
  \adjustbox{max width=\linewidth}{\usebox\FrozenTabBox}}
\makeatletter
\let\ElsParagraph\paragraph
\def\StripRunInStop#1.\StripRunInMark#2\StripRunInEnd{#1}
\let\StripRunInMark\@empty
\renewcommand\paragraph[1]{\ElsParagraph{\StripRunInStop#1\StripRunInMark.\StripRunInMark\StripRunInEnd}}
\newcommand\AlgWrap[1]{\parbox[t]{\dimexpr\linewidth-\ALG@thistlm\relax}{#1\strut}}
\makeatother

\newtheorem{proposition}{Proposition}
\newtheorem{remark}{Remark}
\newtheorem{definition}{Definition}

\newcommand{\E}{\mathbb{E}}
\newcommand{\R}{\mathbb{R}}
\newcommand{\Preal}{P_r}
\newcommand{\Pg}{P_g}
\newcommand{\Lc}{\mathcal{L}}
\newcommand{\norm}[1]{\left\| #1 \right\|}
\newcommand{\abs}[1]{\left| #1 \right|}
\newcommand{\Dtemp}{D_{\mathrm{temp}}}
\newcommand{\Dstat}{D_{\mathrm{stat}}}
\newcommand{\Dsort}{D_{\mathrm{sort}}}
\newcommand{\blambda}{\boldsymbol{\lambda}}
\newcommand{\sort}{\mathrm{sort}}
\newcommand{\argmin}{\operatorname*{arg\,min}}

\newcommand{\SeqSdRcganAnkara}{0.95}
\newcommand{\SeqSdRcganBergen}{0.59}

\newcommand{\SeqSdRealAnkara}{8.21}
\newcommand{\SeqSdRealBergen}{5.18}

\newcommand{\SeqSdRealYakutsk}{21.41}

\newcommand{\SeqSdStatBergen}{4.22}

\newcommand{\SeqSdStatYakutsk}{21.38}

\newcommand{\SeqSdTimeganYakutsk}{19.45}

\newcommand{\WithinSdRealBergen}{2.14}

\newcommand{\WithinSdRealYakutsk}{4.78}

\newcommand{\WithinSdStatBergen}{3.58}

\newcommand{\WithinSdTimeganYakutsk}{9.13}

\newcommand{\ciHiNoCalibrationKs}{0.149}

\newcommand{\ciHiNoDSortedTau}{0.134}

\newcommand{\ciHiRankDSortTau}{0.225}

\newcommand{\ciLoNoCalibrationKs}{0.098}

\newcommand{\ciLoNoDSortedTau}{0.068}

\newcommand{\ciLoRankDSortTau}{0.010}

\newcommand{\ksFloorAnkara}{0.0363}
\newcommand{\ksFloorBergen}{0.0360}
\newcommand{\ksFloorDubai}{0.0344}
\newcommand{\ksFloorLhasa}{0.0846}
\newcommand{\ksFloorYakutsk}{0.0488}

\newcommand{\ksRatioFullMax}{1.04}
\newcommand{\ksRatioFullMin}{1.00}

\newcommand{\ksRatioNoCalAnkara}{5.85}

\newcommand{\ksRatioNoCalLhasa}{2.12}

\newcommand{\medFixedLambdaTau}{-0.002}

\newcommand{\medMomentDStatTau}{-0.022}

\newcommand{\medNoAuxLossesTau}{0.016}

\newcommand{\medNoCalibrationKs}{0.126}
\newcommand{\medNoCalibrationPhys}{-0.010}

\newcommand{\medNoDSortedTau}{0.116}

\newcommand{\medNoDStatTau}{0.032}

\newcommand{\medNoDTemporalTau}{-0.007}
\newcommand{\medRankDSortAcf}{-0.085}

\newcommand{\medRankDSortPhys}{0.142}
\newcommand{\medRankDSortTau}{0.094}

\newcommand{\nLocs}{5}
\newcommand{\nPairs}{25}
\newcommand{\nRunsAbl}{225}

\newcommand{\nSeeds}{5}

\newcommand{\pHolmFixedLambdaTau}{1.000}

\newcommand{\pHolmMomentDStatTau}{0.205}

\newcommand{\pHolmNoAuxLossesTau}{1.000}

\newcommand{\pHolmNoCalibrationKs}{<0.001}
\newcommand{\pHolmNoCalibrationPhys}{<0.001}

\newcommand{\pHolmNoDStatTau}{1.000}

\newcommand{\pHolmNoDTemporalTau}{1.000}
\newcommand{\pHolmRankDSortAcf}{0.025}

\newcommand{\pHolmRankDSortPhys}{<0.001}

\newcommand{\pctNoDSortedTauAnkara}{39}
\newcommand{\pctNoDSortedTauBergen}{16}
\newcommand{\pctNoDSortedTauDubai}{96}
\newcommand{\pctNoDSortedTauLhasa}{13}
\newcommand{\pctNoDSortedTauYakutsk}{82}
\newcommand{\physCalAnkara}{0.0243}

\newcommand{\physCalYakutsk}{0.0778}

\newcommand{\physRawBergen}{0.0056}
\newcommand{\physRawDubai}{0.0001}

\newcommand{\runNoiseBand}{20}
\newcommand{\runNoiseDtau}{0.036}

\newcommand{\verdictDFive}{does not hold}

\newcommand{\codeAdamBetaOne}{0}
\newcommand{\codeAdamBetaTwo}{0.9}
\newcommand{\codeAuxRamp}{15}
\newcommand{\codeAuxStart}{35}
\newcommand{\codeAuxW}{0.01}
\newcommand{\codeDStatHidden}{512}
\newcommand{\codeDStatLayers}{3}
\newcommand{\codeDTempHidden}{256}
\newcommand{\codeDTempLayers}{2}
\newcommand{\codeFitHi}{1}
\newcommand{\codeFitLo}{0.01}
\newcommand{\codeLrDecayDiv}{10}
\newcommand{\codeLrDecayEp}{30}
\newcommand{\codeMomentClamp}{5}
\newcommand{\codeNCritic}{3}
\newcommand{\codeNCriticEp}{50}
\newcommand{\codeNCriticLate}{2}
\newcommand{\codePhysDewW}{50}
\newcommand{\codePhysHumW}{5}
\newcommand{\codePhysPresW}{5}
\newcommand{\codePhysW}{0.1}
\newcommand{\codeQuantLevels}{0.1, 0.25, 0.5, 0.75, 0.9}
\newcommand{\codeRankAnchors}{32}
\newcommand{\codeRankTemp}{0.05}
\newcommand{\codeTailFracMax}{0.3}
\newcommand{\codeTailScale}{3}
\newcommand{\codeTailStart}{30}
\newcommand{\codeTendLag}{3}
\newcommand{\codeTgDThresh}{0.15}
\newcommand{\codeTgSupLayers}{2}
\newcommand{\dKsMaxLhasa}{0.0110}
\newcommand{\dKsMedianAll}{0.0000}
\newcommand{\ksRatioFullMedian}{1.019}
\newcommand{\nBootDistinct}{126}
\newcommand{\nHolmSigWord}{six}
\newcommand{\nLocsProtoWord}{five}
\newcommand{\nPosSeedsNoDSortedTauBergen}{3}
\newcommand{\nPosSeedsNoDSortedTauLhasa}{3}
\newcommand{\nProjRuns}{25}
\newcommand{\pHolmBaseRcganKs}{=0.420}
\newcommand{\pHolmBaseRcganPhys}{<0.001}
\newcommand{\pHolmBaseRcganTau}{<0.001}
\newcommand{\pHolmBaseTimeganAcf}{<0.001}
\newcommand{\pHolmBaseTimeganKs}{=0.634}
\newcommand{\pHolmBaseTimeganTau}{=0.002}
\newcommand{\pHolmRelNoDSortedTau}{=0.002}
\newcommand{\pHolmRelRankDSortTau}{=0.006}
\newcommand{\phDKsLocsOverThrList}{Bergen, Dubai and Lhasa}
\newcommand{\phDKsLocsOverThrWord}{three}
\newcommand{\phDistBelowAll}{59}
\newcommand{\phDistBelowMaxRatio}{0.562}
\newcommand{\phDistDevice}{NVIDIA RTX PRO 6000 Blackwell Server Edition}
\newcommand{\phDistGapMed}{0.023}
\newcommand{\phDistGapOverDMed}{0.195}
\newcommand{\phDistGateMaxDiff}{0}
\newcommand{\phDistGateTol}{10^{-4}}
\newcommand{\phDistNRuns}{250}
\newcommand{\phDistRatioMed}{0.134}
\newcommand{\phDistRatioPninety}{0.292}
\newcommand{\phDistTopOneLoc}{Bergen}
\newcommand{\phDistTopOneModel}{TimeGAN}
\newcommand{\phDistTopOneRatio}{0.699}
\newcommand{\phDistTopOneSeed}{42}
\newcommand{\phDistTopThreeLoc}{Bergen}
\newcommand{\phDistTopThreeModel}{\texttt{no\_D\_sorted}}
\newcommand{\phDistTopThreeRatio}{0.475}
\newcommand{\phDistTopThreeSeed}{2024}
\newcommand{\phDistTopTwoLoc}{Dubai}
\newcommand{\phDistTopTwoModel}{\texttt{no\_D\_sorted}}
\newcommand{\phDistTopTwoRatio}{0.512}
\newcommand{\phDistTopTwoSeed}{2024}
\newcommand{\phDistTorch}{2.11.0+cu130}
\newcommand{\phFitLastEpochs}{100}
\newcommand{\phFitSorted}{1.0}
\newcommand{\phFitStat}{36.6}
\newcommand{\phFitTemporal}{26.1}
\newcommand{\phKsRatioCalFullMed}{1.02}
\newcommand{\phKsRatioCalRcganMed}{1.00}
\newcommand{\phKsRatioCalTimeganMed}{1.00}
\newcommand{\phKsRatioCells}{200}
\newcommand{\phKsRatioCellsBelowOne}{39}
\newcommand{\phKsRatioRawFullMed}{4.28}
\newcommand{\phKsRatioRawRcganMed}{11.66}
\newcommand{\phKsRatioRawTimeganMed}{2.08}
\newcommand{\phLambdaSortAtFloorPct}{87}
\newcommand{\phNRunsLambda}{25}
\newcommand{\phRatioCapBindsPct}{65}
\newcommand{\phSeedsBelowTenRcgan}{8}
\newcommand{\phSeedsPerModel}{25}
\newcommand{\phTrainTorch}{2.11.0+cu128}
\newcommand{\protoAcfLags}{24}
\newcommand{\protoAcfNSeq}{200}
\newcommand{\protoBatch}{512}
\newcommand{\protoBootCI}{95}
\newcommand{\protoDFiveFrac}{0.05}
\newcommand{\protoDSixPct}{10}
\newcommand{\protoDewTol}{0.1}
\newcommand{\protoEmbargoHours}{168}
\newcommand{\protoEpochs}{150}
\newcommand{\protoEvoEta}{0.03}
\newcommand{\protoEvoHi}{6}
\newcommand{\protoEvoLo}{0.5}
\newcommand{\protoEvoRatio}{3}
\newcommand{\protoGpFreq}{5}
\newcommand{\protoGpLambda}{10}
\newcommand{\protoHeldOutFirst}{2023}
\newcommand{\protoHidden}{512}
\newcommand{\protoHumHi}{100}
\newcommand{\protoHumLo}{0}
\newcommand{\protoLatent}{128}
\newcommand{\protoLrG}{10^{-3}}
\newcommand{\protoNAcfFeatures}{5}
\newcommand{\protoNAcfFeaturesWord}{five}
\newcommand{\protoNArms}{9}
\newcommand{\protoNBaselinesWord}{two}
\newcommand{\protoNBoot}{20{,}000}
\newcommand{\protoNContrastsWord}{eight}
\newcommand{\protoNEval}{5{,}000}
\newcommand{\protoNFeatures}{11}
\newcommand{\protoNQuantiles}{1{,}000}
\newcommand{\protoNRecordHours}{131{,}496}
\newcommand{\protoNScaled}{7}
\newcommand{\protoNScaledWord}{seven}
\newcommand{\protoNSeedsWord}{five}
\newcommand{\protoNTrainSeq}{4{,}735}
\newcommand{\protoNValHours}{17{,}544}
\newcommand{\protoNValSeq}{725}
\newcommand{\protoPresHi}{1045}
\newcommand{\protoPresLo}{980}
\newcommand{\protoRcEpochs}{100}
\newcommand{\protoRcHidden}{128}
\newcommand{\protoRcLatent}{32}
\newcommand{\protoRcLayers}{2}
\newcommand{\protoRcLr}{10^{-3}}
\newcommand{\protoSeqLen}{168}
\newcommand{\protoSignTestPMin}{0.0625}
\newcommand{\protoStride}{24}
\newcommand{\protoTauSubsample}{50{,}000}
\newcommand{\protoTgHidden}{64}
\newcommand{\protoTgItersAe}{2{,}000}
\newcommand{\protoTgItersJoint}{2{,}000}
\newcommand{\protoTgItersSup}{2{,}000}
\newcommand{\protoTgLayers}{3}
\newcommand{\protoValYears}{2}
\newcommand{\protoValYearsWord}{two}
\newcommand{\protoWarmup}{15}
\newcommand{\protoYearFirst}{2010}
\newcommand{\protoYearLast}{2024}

\journal{Computational Statistics and Data Analysis}

\begin{document}

\begin{frontmatter}

\title{Marginal Calibration Masks Generator Differences in Held-Out
       Evaluation of Multivariate Time Series, with an Application to
       ERA5 Weather Sequences\tnoteref{t1}}

\tnotetext[t1]{This paper is derived from the first author's Master's
thesis, ``Game Theory Based Adversarial Learning for Synthetic
Multivariate Data Generation: A Novel GAN Architecture
(StatD\textsuperscript{2}GAN),'' Department of Statistics, Graduate
School of Science and Engineering, Hacettepe University, 2026.}

\author[inst1]{Mustafa Özaytaç\corref{cor1}\fnref{orcid1}}
\ead{mustafaozaytac@hacettepe.edu.tr}
\cortext[cor1]{Corresponding author.}
\fntext[orcid1]{ORCID: 0009-0009-8486-0828}

\author[inst2]{\"{O}zge Karada\u{g} Ata\c{s}\fnref{orcid2}}
\ead{ozgekaradag@hacettepe.edu.tr}
\fntext[orcid2]{ORCID: 0000-0002-2650-1458}

\affiliation[inst1]{organization={Department of Statistics, Graduate School of Science and Engineering, Hacettepe University},
             city={Ankara},
             postcode={06800},
             country={Turkey}}

\affiliation[inst2]{organization={Department of Statistics, Faculty of Science, Hacettepe University},
             city={Ankara},
             postcode={06800},
             country={Turkey}}

\begin{abstract}
Generative models for multivariate time series are routinely
evaluated with pooled distributional metrics computed after marginal
calibration. We show that, under a held-out protocol, the calibrated
Kolmogorov--Smirnov distance differs from a real-versus-real floor by at
most the distance of the calibrated sample to the training block, so that
it carries little information about the generator, and we audit a
three-discriminator GAN with evolutionary weight adaptation (StatD2GAN)
against re-implemented RCGAN and TimeGAN baselines under a pre-specified
protocol on ERA5 weather sequences: the final \protoValYearsWord{} calendar years of each
dataset are held out behind a \protoEmbargoHours{} hour embargo,
calibration is fitted on the training block only, and the decision rules
were pre-specified. Evidence comes from \nPairs{} matched (location,
seed) pairs across \nLocsProtoWord{} locations, tested with
Holm-corrected Wilcoxon signed-rank tests. First, isotonic calibration
drives the Kolmogorov--Smirnov distance to a median of
\ksRatioFullMedian{} times a per-location noise-and-shift floor (the
bound confines it to within \phDistRatioMed{} times the floor at the
median), and
neither baseline is distinguishable from StatD2GAN on it, so this calibrated KS metric does not discriminate among the evaluated
generators at this resolution. Second, the
sorted-representation discriminator is the only component whose removal
significantly degrades cross-variable dependence (Kendall $\tau$ MAE
$+\medNoDSortedTau$, $p_{\mathrm{Holm}}\pHolmRelNoDSortedTau$). A
rank-transformed variant also degrades Kendall $\tau$ MAE, consistent
with quantile supervision rather than copula matching. Third, physical
constraint violations are raised by calibration rather than by the
generator; projection removes them, although the largest marginal change
(\dKsMaxLhasa{}, Lhasa) exceeds the pre-specified tolerance in
\phDKsLocsOverThrWord{} locations. Fourth, the
re-implemented TimeGAN is significantly better than StatD2GAN on Kendall
$\tau$ and ACF MAE, RCGAN is worse on Kendall $\tau$ MAE and constraint
violations, and the pre-specified location-level collapse criterion was
not met, although RCGAN fell below the threshold in
\phSeedsBelowTenRcgan{} of \phSeedsPerModel{} runs. We recommend
floor-referenced marginal evaluation, matched-pair testing, and
dependence and sequence-level metrics.
\end{abstract}


\end{frontmatter}


\section{Introduction}
\label{sec:intro}

Evaluating generators of multivariate time series is itself a
statistical problem, and generating high-fidelity synthetic series is a
central challenge in domains where data are scarce, sensitive, or costly
to obtain.  For meteorological applications, the utility of synthetic data
hinges on two properties that are often in tension: \textit{statistical
fidelity}, meaning the generated distribution should match the real
data's marginal and joint structure, and \textit{physical plausibility},
meaning generated sequences must respect the bound and range constraints
(dewpoint-temperature ordering, pressure and humidity ranges) that define
domain-admissible states.

Generative Adversarial Networks (GANs) \cite{goodfellow2014gan} are a
dominant paradigm for this task, and a common pipeline pairs a GAN with a
post-hoc marginal calibration step, such as per-feature isotonic
regression, to close the residual gap between generated and real
marginals.  This paper starts from a methodological observation about
that pipeline.  If the calibration map is fitted on the same data on
which distributional metrics are subsequently computed, the evaluation is
circular: the metric certifies that calibration ran, not that the
generator learnt anything.  We report the study in full, including
the pre-specified predictions that were not borne out.

The vehicle is \textbf{StatD2GAN}, a three-discriminator architecture
that decomposes the adversarial objective into temporal realism
($\Dtemp$), statistical fidelity ($\Dstat$), and sorted-representation
structure ($\Dsort$), trained under WGAN-GP \cite{gulrajani2017improved}
with fitness-proportionate replicator dynamics
\cite{taylor1978evolutionary} adapting the discriminator weights, and
completed by isotonic calibration \cite{barlow1972isotonic} and a
physical constraint projection.  We evaluate it against re-implemented
RCGAN \cite{esteban2017rcgan} and TimeGAN \cite{yoon2019timegan}
baselines on \nLocsProtoWord{} ERA5~\cite{hersbach2020era5} locations in contrasting climates,
under a held-out protocol: the final
\protoValYearsWord{} calendar years of each series are reserved for
evaluation behind a \protoEmbargoHours{} hour embargo, calibration is
fitted on the training block only, and every metric is computed on the
held-out block.  The protocol and the decision rules for the principal
claims were fixed and hashed before the evaluation campaign.  All claims
rest on \nPairs{} matched (location, seed) pairs tested with Wilcoxon
signed-rank tests \cite{wilcoxon1945} under Holm correction
\cite{holm1979}; observed run-to-run noise (median
$\abs{\Delta\tau} = \runNoiseDtau$ between identically configured runs)
rules out single-seed comparisons.  We make no claim that StatD2GAN is
superior to existing generators.

\medskip
\noindent The contributions of this work are:
\begin{enumerate}
  \item An argument that circular evaluation in calibrated generative
        pipelines can invalidate architectural conclusions, and a held-out
        protocol (temporal split, embargo, train-fitted calibration) that
        removes it.  The study contains no circular-calibration arm; the
        evidence is the behaviour of calibrated metrics under the held-out
        protocol, where calibration drives the Kolmogorov--Smirnov distance to a median
        of \ksRatioFullMedian{} times a per-location real-versus-real
        noise-and-shift floor (location means \ksRatioFullMin{} to
        \ksRatioFullMax), as the map is built to do, and neither baseline, including the under-performing RCGAN, is
        distinguishable from
        StatD2GAN on it ($p_{\mathrm{Holm}}\pHolmBaseRcganKs$ and
        $p_{\mathrm{Holm}}\pHolmBaseTimeganKs$): calibrated marginal metrics carry no detectable architectural information
        at this resolution.
  \item A floor-referenced formulation of marginal evaluation, in
        which the KS sup-distance is reported relative to the resampling
        noise-and-shift floor of the real held-out data, replacing threshold-based
        pass rates.
  \item A mechanism dissection of the sorted-representation
        discriminator, the only component whose removal significantly
        degrades dependence structure (Kendall $\tau$ MAE
        $+\medNoDSortedTau$, $p_{\mathrm{Holm}}\pHolmRelNoDSortedTau$).
        The location-mean increase ranges from \pctNoDSortedTauLhasa\%
        (Lhasa) to \pctNoDSortedTauDubai\% (Dubai); in Bergen and Lhasa it
        lies within the $\pm$\runNoiseBand\% run-to-run band, with
        \nPosSeedsNoDSortedTauBergen{} and \nPosSeedsNoDSortedTauLhasa{}
        of \nSeeds{} seeds positive.  A rank-transformed ablation arm also
        degrades Kendall $\tau$ MAE ($+\medRankDSortTau$,
        $p_{\mathrm{Holm}}\pHolmRelRankDSortTau$), which is consistent with
        the channel supervising marginal quantiles rather than copula
        structure.  All remaining
        components, including the temporal and statistical discriminators,
        the moment-based variant, the auxiliary losses and the
        evolutionary weighting, are statistically
        indistinguishable from the full model after correction, and we
        report these null results explicitly.
  \item Evidence that physical constraint violations are raised by the
        marginal-independent calibration map rather than by the generator
        (raw violation rates at most \physRawBergen{} against
        \physCalAnkara{} to \physCalYakutsk{} after calibration), and a
        constraint projection step that eliminates them in all
        \nProjRuns{} full-model runs at a small, run-dependent
        distributional cost: the median change in KS is \dKsMedianAll{} and
        the maximum \dKsMaxLhasa, but the pre-specified tolerance,
        \protoDFiveFrac{} times the mean reference distance across locations,
        is exceeded in \phDKsLocsOverThrWord{} locations
        (\phDKsLocsOverThrList; post-freeze descriptive), so this
        prediction was not borne out.
  \item A demonstration that the calibrated marginal metric does not
        rank generators that dependence metrics separate: TimeGAN is
        significantly better than StatD2GAN on Kendall $\tau$ MAE
        ($p_{\mathrm{Holm}}\pHolmBaseTimeganTau$) and ACF MAE
        ($p_{\mathrm{Holm}}\pHolmBaseTimeganAcf$), and RCGAN is worse on
        Kendall $\tau$ MAE ($p_{\mathrm{Holm}}\pHolmBaseRcganTau$) and
        constraint violations ($p_{\mathrm{Holm}}\pHolmBaseRcganPhys$).
        The pre-specified location-level collapse criterion for
        between-sequence weekly-mean variability was not met, although
        RCGAN fell below the \protoDSixPct\% threshold in
        \phSeedsBelowTenRcgan{} of \phSeedsPerModel{} runs (post-freeze
        descriptive).
\end{enumerate}

The remainder of the paper is organised as follows.
Section~\ref{sec:related} reviews related work.
Section~\ref{sec:method} specifies the architecture, losses, and the
calibration and projection stages.  Section~\ref{sec:setup} defines the
evaluation protocol, metrics, and statistical methodology.
Section~\ref{sec:results} presents the results,
Section~\ref{sec:discussion} discusses them, and
Section~\ref{sec:conclusion} concludes.

\section{Related Work}
\label{sec:related}

\subsection{Multi-Discriminator GANs}

Nguyen et al.~\cite{nguyen2017d2gan} introduced D2GAN with two discriminators
whose opposing loss directions improve mode coverage relative to the
vanilla GAN.  Chandana et al.~\cite{kurri2024generalised} unified dual
discriminator objectives under a generalised $\alpha$-loss framework and
showed that the resulting min-max problem reduces to minimising a linear
combination of an $f$-divergence and its reverse.  MCLGAN \cite{mclgan2021}
introduced a specialisation strategy for multi-discriminator training
through competitive learning, assigning distinct data regions to individual
discriminators.  DUEL \cite{duelgan2022} extended the inter-discriminator
duel mechanism to encourage diversity.  The game-theoretic grounding of
multi-discriminator stability, particularly Nash equilibrium analysis and
convergence conditions, is discussed in \cite{mescheder2018convergence}.

\subsection{Time Series GANs}

TimeGAN \cite{yoon2019timegan} introduced an embedding network for joint
training in latent and feature space with a supervised loss term for
temporal dynamics.  RCGAN \cite{esteban2017rcgan} proposed conditional
recurrent generation for medical time series.  SeriesGAN \cite{seriesgan}
focused on autocorrelation structure preservation.  The formal problem of
constrained time series generation, enforcing structural constraints
without post-hoc repair, is studied in \cite{lin2020constrained}.
Architectural refinements to the Long Short-Term Memory (LSTM) backbone itself, such as
attention-based hidden-state adjustment within a GAN generator and
discriminator, have also been explored for sequence modelling
\cite{adjustedlstm}.

\subsection{Physics-Constrained GANs}

Physically constrained GANs embed domain knowledge as differentiable
soft or hard penalties in the generator's loss
\cite{physics_gan_turb,physics_constrained_gan}.  The PINT framework
\cite{pint} formalised physics-informed training for neural generators.
Applications to turbulence simulation~\cite{physics_gan_turb} and to
precipitation fields from Earth system models~\cite{physics_constrained_gan}
use physics-aware loss terms to reduce constraint violations.
Our domain-constrained loss follows the soft-penalty formulation of
\cite{physics_constrained_gan}, adapted to ERA5 meteorological variables;
unlike the PDE- and thermodynamic-closure-based penalties surveyed above,
our three constraints are variable bounds and range checks (dewpoint
below temperature, valid pressure and humidity ranges), not physical
conservation laws.

\medskip
Evaluation methodology for generative time series models has received
less attention than architecture design.  Threshold-based marginal tests
and pooled dependence metrics are standard, but their behaviour under
post-hoc calibration, which by construction targets exactly the marginal
statistics they measure, has not to our knowledge been analysed.  The
protocol contribution of this paper addresses that gap.

\section{Methodology}
\label{sec:method}

\subsection{Problem Formulation}

Let $\mathcal{X} \subset \R^{T \times F}$ denote the space of multivariate
time series of length $T$ with $F$ features.  Let $\Preal$ be the unknown
real data distribution and $\mathbf{x} \sim \Preal$ denote real
observations.  The generator $G: \R^{d_z} \to \R^{T \times F}$ maps latent
noise $\mathbf{z} \sim \mathcal{N}(\mathbf{0}, \mathbf{I}_{d_z})$ to
synthetic sequences, inducing $\Pg$ defined by
$\tilde{\mathbf{x}} = G(\mathbf{z})$.  The generator is a three-layer LSTM
whose output layer is spectrally normalised; the two (sine, cosine)
feature pairs are projected onto the unit circle at the output.

\subsection{Multi-Discriminator Adversarial Objective}

\paragraph{WGAN-GP Foundation.}
We adopt the Wasserstein-1 distance
\cite{arjovsky2017wasserstein} via the Kantorovich--Rubinstein duality
\cite{villani2009optimal}:
\begin{equation}
  W(\Preal, \Pg)
    = \sup_{\norm{f}_L \leq 1}
      \E_{\mathbf{x}\sim\Preal}\!\left[f(\mathbf{x})\right]
      - \E_{\tilde{\mathbf{x}}\sim\Pg}\!\left[f(\tilde{\mathbf{x}})\right]
  \label{eq:wasserstein}
\end{equation}
Lipschitz control is applied through two complementary mechanisms.  All
discriminator (and generator output) linear and convolutional layers use
spectral normalisation, and the gradient penalty
\cite{gulrajani2017improved}
\begin{equation}
  \Lc_{\mathrm{GP}}(D)
    = \E_{\hat{\mathbf{x}}\sim P_{\hat{x}}}\!
      \left[\left(
        \norm{\nabla_{\hat{\mathbf{x}}} D(\hat{\mathbf{x}})}_2 - 1
      \right)^{\!2}\right]
  \label{eq:gp}
\end{equation}
with $\hat{\mathbf{x}} = \alpha\,\mathbf{x} + (1-\alpha)\,\tilde{\mathbf{x}}$,
$\alpha \sim \mathcal{U}[0,1]$, is evaluated lazily on every \protoGpFreq{}th batch
with its coefficient rescaled by the same factor
($\lambda_{\mathrm{GP}} = \protoGpLambda$ in expectation), so that the
average regularisation pressure matches a per-batch penalty at a
fraction of the cost.

\paragraph{Objective.}
StatD2GAN couples one generator with three critics:
\begin{equation}
  \min_G \; \max_{\Dtemp,\,\Dstat,\,\Dsort} \;
    \sum_{k \in \{\mathrm{temp},\mathrm{stat},\mathrm{sort}\}}
      \lambda_k \Lc_{D_k}
    + \lambda_{\mathrm{GP}} \Lc_{\mathrm{GP}}
  \label{eq:main_obj}
\end{equation}
where each critic loss takes the standard WGAN form~\eqref{eq:wasserstein}.
The generator minimises
\begin{equation}
  \Lc_G
    = \sum_k \lambda_k\,
      \E_\mathbf{z}\!\left[-D_k(G(\mathbf{z}))\right]
    + \lambda_{\mathrm{phys}} \Lc_{\mathrm{phys}}(G)
    + \Lc_{\mathrm{aux}}(G)
  \label{eq:gen_loss}
\end{equation}
with $\lambda_{\mathrm{phys}} = \codePhysW$.
$\Lc_{\mathrm{aux}}$ collects two distribution-matching auxiliary losses
defined in Section~\ref{sec:training}.

\subsection{Discriminator Specifications}
\label{sec:discriminators}

\paragraph{$\Dtemp$: Temporal Discriminator.}
$\Dtemp$ is a \codeDTempLayers-layer LSTM (hidden size \codeDTempHidden)
reading the raw sequence;
the final hidden state feeds a spectrally normalised linear head:
\begin{equation}
  \mathbf{h}_t = \mathrm{LSTM}\!\left(\mathbf{x}_t,\, \mathbf{h}_{t-1}\right),
  \quad
  \Dtemp(\mathbf{x}) = \mathbf{w}^\top \mathbf{h}_T + b
  \label{eq:dtemp}
\end{equation}
It is sensitive to temporal transition dynamics and sequential
autocorrelation structure.

\paragraph{$\Dstat$: Statistical Fidelity Discriminator.}
$\Dstat$ is a deeper recurrent critic, a \codeDStatLayers-layer LSTM (hidden size
\codeDStatHidden) with a spectrally normalised linear head, intended to capture
distributional properties that the shallower $\Dtemp$ misses through
greater capacity rather than through hand-crafted features.  An
alternative variant that replaces the LSTM with a network operating on a
per-feature moment summary vector (mean, variance, skewness, excess
kurtosis) is evaluated as the \texttt{moment\_D\_stat} ablation
arm\footnote{In the moment variant, moments are computed in single
precision outside mixed-precision autocast, standardised deviations are
clamped to $[-\codeMomentClamp, \codeMomentClamp]$, and the moment
vector is layer-normalised.
Without these safeguards, fourth-moment gradients explode under the
tail-widened latent sampling of Section~\ref{sec:training}.}; it is
statistically indistinguishable from the LSTM form on every metric
(Section~\ref{sec:results_ablation}), so the extra inductive bias gives
no detectable benefit here.

\paragraph{$\Dsort$: Sorted-Representation Discriminator.}
\begin{definition}[Sorted Representation]
For $\mathbf{x} \in \R^{T\times F}$, the sorted representation
$\check{\mathbf{x}} \in \R^{T\times F}$ is obtained by independently
sorting each feature column in ascending order:
\begin{equation}
  \check{x}_{:,f} = \sort\!\left(x_{:,f}\right), \quad f = 1,\ldots,F
  \label{eq:sorted}
\end{equation}
\end{definition}
Column-wise sorting maps each feature to its within-sequence empirical
quantile function.  The representation therefore preserves the full
empirical marginal of every feature.  Because all columns are permuted
by their own ordering, the original cross-variable pairing at each time
step is destroyed along with the temporal ordering; what remains across
columns is the alignment of quantile levels.  $\Dsort$ applies two
spectrally normalised circular convolutions
\begin{equation}
  (\check{\mathbf{x}} \circledast w)[i]
    = \sum_{k=0}^{K-1} w[k]\cdot \check{x}[(i-k) \bmod T]
  \label{eq:circ_conv}
\end{equation}
followed by global max pooling and a linear head.

\begin{remark}
It is tempting to read the sorted representation as retaining the
cross-variable rank (copula \cite{nelsen2006copulas}) structure.  The
ablation evidence in Section~\ref{sec:results_dissection} does not
support that reading: a diagnostic variant (\texttt{rank\_D\_sort},
below) that carries the copula and also keeps the temporal order, but
not the quantile values, does not recover the effect.  The evidence is
consistent with $\Dsort$ supervising the per-feature quantile functions,
with an indirect downstream effect on dependence metrics.
\end{remark}

\paragraph{Rank-transformed variant (mechanism probe).}
To separate quantile supervision from copula matching, we construct
\texttt{rank\_D\_sort}: the same convolutional body as $\Dsort$, but fed
a differentiable rank transform instead of the column-wise sort.  Each
entry is mapped to its within-sequence empirical CDF value through a
sigmoid-smoothed comparison against \codeRankAnchors{} deterministic
quantile anchors (temperature $\codeRankTemp$; anchors detached), rescaled to $[-1,1]$.  Unlike
sorting, this transform preserves the row pairing across features and the
temporal order, so
its output carries the empirical copula by construction.  If the benefit
of $\Dsort$ flowed through copula information, this variant should
reproduce or exceed it; Section~\ref{sec:results_dissection} shows it
does not.

\subsection{Domain-Constrained Loss}
\label{sec:physics}

Domain constraints are embedded as differentiable soft penalties on the
generated sequence in physical units.  Three constraints are enforced,
with weights $(\alpha_1, \alpha_2, \alpha_3) =
(\codePhysDewW, \codePhysPresW, \codePhysHumW)$:
\begin{align}
  c_1:\;& T_{\mathrm{dew},t} \leq T_{2m,t} + \protoDewTol\,\mathrm{K}
    \tag{C1: Dewpoint bound} \label{eq:c1}\\
  c_2:\;& \protoPresLo \leq P_{\mathrm{msl},t} \leq \protoPresHi \;\mathrm{hPa}
    \tag{C2: Pressure band} \label{eq:c2}\\
  c_3:\;& \protoHumLo \leq \mathrm{RH}_t \leq \protoHumHi\,\%
    \tag{C3: Humidity bounds} \label{eq:c3}
\end{align}
The dewpoint term uses a squared softplus for smooth gradients near the
boundary; the band constraints use rectified linear penalties on both
sides.  The total
$\Lc_{\mathrm{phys}}$ enters the generator loss with the global factor
$\lambda_{\mathrm{phys}} = \codePhysW$ in Eq.~\eqref{eq:gen_loss}.
The constraint thresholds are defined once and shared verbatim by the
training penalty, the violation metric of Section~\ref{sec:metrics}, and
the projection step of Section~\ref{sec:calibration}, so no
train-versus-evaluation constant mismatch is possible.  Constraints C1
to C3 define the domain-admissible subspace
$\mathcal{X}_{\mathrm{phys}} \subset \mathcal{X}$ (subscript retained for
consistency with the code and metric names below).

\subsection{Evolutionary Weight Adaptation}
\label{sec:evo}

The critic weights $\blambda = [\lambda_1, \lambda_2, \lambda_3]$ are
initialised uniformly, $\lambda_k^{(0)} = 1$, and adapted during
training by a discrete replicator rule \cite{taylor1978evolutionary}.
Let $\ell_k^{(t)}$ be the magnitude of critic $k$'s Wasserstein estimate
on the current batch, and
\begin{equation}
  f_k^{(t)} = \mathrm{clip}\!\left(
    \frac{\ell_k^{(t)}}{\sum_j \ell_j^{(t)} + \epsilon},\; \codeFitLo,\; \codeFitHi
  \right),
  \qquad
  \bar{f}^{(t)}
    = \frac{\sum_k \lambda_k^{(t)} f_k^{(t)}}{\sum_k \lambda_k^{(t)}}
  \label{eq:fitness}
\end{equation}
the normalised fitness and its weight-averaged mean.  The update is
\begin{equation}
  \lambda_k^{(t+1)}
    = \lambda_k^{(t)}
      \cdot \left[1 + \eta \left(f_k^{(t)} - \bar{f}^{(t)}\right)\right]
  \label{eq:replicator}
\end{equation}
with $\eta = \protoEvoEta$, followed by three safeguards: clipping to
$[\lambda_{\min}, \lambda_{\max}] = [\protoEvoLo, \protoEvoHi]$, capping the ratio
$\max_k \lambda_k / \min_k \lambda_k$ at $\protoEvoRatio$, and renormalising
$\sum_k \lambda_k$ to its initial value.

One property of this scheme should be stated plainly.  The raw fitness
magnitudes of the three critics differ by more than an order of
magnitude (median logged values
$\hat{W} \approx [\phFitTemporal,\, \phFitStat,\, \phFitSorted]$ for
$\Dtemp$, $\Dstat$, $\Dsort$ over the final \phFitLastEpochs{} epochs of
the \phNRunsLambda{} full-model runs), so after normalisation the
adaptation is driven almost entirely by the two recurrent critics.  In
the median run the ratio cap binds in \phRatioCapBindsPct\% of epochs and
the weight of $\Dsort$ sits at its lower bound in
\phLambdaSortAtFloorPct\%.  The replicator mechanism is best understood
as a bounded rebalancing heuristic, not as a solution concept; we do not
attach a game-theoretic optimality claim to it, and its empirical
contribution is evaluated as the \texttt{fixed\_lambda} ablation arm.

\subsection{Two-Phase Training and Auxiliary Losses}
\label{sec:training}

Training proceeds in two phases (Algorithm~\ref{alg:training}).

\paragraph{Phase 1: Distributional warmup.}
For the first \protoWarmup{} epochs the generator is trained without adversarial
feedback, minimising a sorted-sample CDF matching loss, a quantile
matching loss at the $\{\codeQuantLevels\}$ levels, and
$\Lc_{\mathrm{phys}}$.  This anchors the marginals before the critics
begin shaping the sequence structure.

\paragraph{Phase 2: Adversarial training.}
From epoch \protoWarmup{} the full objective~\eqref{eq:main_obj} is
optimised with $n_{\mathrm{critic}} = \codeNCritic$ critic steps per
generator step (reduced to \codeNCriticLate{} after epoch \codeNCriticEp)
and the generator learning rate decayed by a factor of \codeLrDecayDiv{}
at epoch \codeLrDecayEp.  From epoch \codeAuxStart{} the two warmup
losses re-enter the generator objective as auxiliary terms
$\Lc_{\mathrm{aux}}$, with a weight ramped linearly from zero to
\codeAuxW{} over \codeAuxRamp{} epochs.  Latent sampling uses progressive
tail widening: after epoch \codeTailStart{} a growing fraction of each
batch (up to \codeTailFracMax) is drawn at \codeTailScale{} times the
standard deviation,
encouraging coverage of distribution tails.  The contribution of the auxiliary losses is evaluated as the
\texttt{no\_aux\_losses} ablation arm, which removes both the warmup
phase and the auxiliary terms.

\begin{algorithm}[t]
\caption{StatD2GAN training}
\label{alg:training}
\begin{algorithmic}[1]
\Require $\lambda_k^{(0)} = 1$, $\eta = \protoEvoEta$,
         $[\lambda_{\min},\lambda_{\max}] = [\protoEvoLo, \protoEvoHi]$,
         ratio cap $\protoEvoRatio$,
         $\lambda_{\mathrm{GP}} = \protoGpLambda$ (lazy, every
         \protoGpFreq{}th batch), $\lambda_{\mathrm{phys}} = \codePhysW$
\For{each of the first \protoWarmup{} epochs}  \Comment{Phase 1: warmup}
    \State Update $G$ on
           $\Lc_{\mathrm{CDF}} + \Lc_{\mathrm{quant}} + \Lc_{\mathrm{phys}}$
\EndFor
\For{each remaining epoch up to \protoEpochs}  \Comment{Phase 2: adversarial}
    \For{each batch}
        \For{$n_{\mathrm{critic}}$ steps}
            \State \AlgWrap{Update each $D_k$ on its WGAN loss;
                   add $\Lc_{\mathrm{GP}}$ on every fifth batch}
        \EndFor
        \State Update $\blambda$ via
               Eqs.~\eqref{eq:fitness}--\eqref{eq:replicator} and safeguards
        \State Update $G$ on Eq.~\eqref{eq:gen_loss}
               (auxiliary terms active from epoch \codeAuxStart)
    \EndFor
\EndFor
\State \textbf{Calibrate} on the training block
       (Section~\ref{sec:calibration}), \textbf{project} onto
       $\mathcal{X}_{\mathrm{phys}}$
\end{algorithmic}
\end{algorithm}

\subsection{Isotonic Calibration and Constraint Projection}
\label{sec:calibration}

\paragraph{Calibration.}
After training, per-feature isotonic regression
\cite{barlow1972isotonic} closes the residual gap between generated and
real marginals.  For feature $f$, the map $\phi_f$ solves
\begin{equation}
  \phi_f^*
    = \argmin_{\phi\,\in\,\mathcal{M}}
      \sum_{\alpha \in \mathcal{A}}
      \left(\phi\!\left(q_\alpha^{\mathrm{syn}}\right)
            - q_\alpha^{\mathrm{real,train}}\right)^{\!2}
  \label{eq:isotonic}
\end{equation}
where $\mathcal{M}$ is the set of non-decreasing maps and $\mathcal{A}$
is a grid of \protoNQuantiles{} equally spaced quantile levels.  In
practice the minimiser interpolates the quantile pairs
(Section~\ref{sec:results_floor}).  Critically for the
evaluation protocol of Section~\ref{sec:protocol}, the real quantiles
$q_\alpha^{\mathrm{real,train}}$ are computed on the training block
only; the held-out block never touches the calibration fit.  Monotone
maps preserve within-feature rank order, but because each $\phi_f$ acts
independently and may contain flat segments, cross-feature relationships
at a given time step are not protected: Section~\ref{sec:results_cal}
shows that physical constraint violations in the pipeline are raised
by calibration rather than by the generator.

\paragraph{Constraint projection.}
The final pipeline stage projects each calibrated sequence onto
$\mathcal{X}_{\mathrm{phys}}$ by minimal per-constraint corrections:
dewpoint is clipped to the concurrent temperature plus the
\protoDewTol\,K tolerance, humidity to $[\protoHumLo, \protoHumHi]$, and
pressure to its band.  The
projection touches only constraint-violating entries.  Its distributional cost is measured directly in
Section~\ref{sec:results_cal}; the pipeline includes it as its final
stage.

\section{Experimental Setup}
\label{sec:setup}

\subsection{Evaluation Protocol}
\label{sec:protocol}

The central methodological requirement of this study is that no
information from the evaluation data may influence any fitted component
of the pipeline, including the calibration map.  The protocol has three
elements.

\paragraph{Temporal held-out split.}
For each location, the final two full calendar years of the hourly
record are reserved as the held-out evaluation block.  Using complete
calendar years guarantees that every season appears in the evaluation
data, so seasonal coverage cannot differ between configurations.

\paragraph{Embargo.}
A gap of \protoEmbargoHours{} hours, equal to the sequence length, is left between the
end of the training block and the start of the evaluation block, and
sequences overlapping the gap are discarded.  Without the embargo, a
training sequence and an evaluation sequence could share up to $\protoSeqLen - 1$ hourly observations, leaking evaluation data into training.

\paragraph{Train-fitted calibration.}
The isotonic calibration map of Section~\ref{sec:calibration} is fitted
exclusively against training-block quantiles and then applied to
generated data, using a synthetic sample drawn independently of the
one that is evaluated; all metrics are computed against the held-out
block.  Under a circular design, with calibration fitted on the data
used for evaluation, the calibrated marginal metric would measure the
identity map plus resampling noise.

\subsection{Datasets}

We use ERA5 reanalysis data from the Copernicus Climate Data Store
across \nLocsProtoWord{} locations in contrasting climates
(Table~\ref{tab:datasets}).  Default hyperparameters were fixed during
development of the generator (\ref{app:sensitivity}) and are shared by all
locations. Each record spans \protoYearFirst{} to \protoYearLast{}
(\protoNRecordHours{} hours); the held-out block is \protoHeldOutFirst{} to
\protoYearLast.

\begin{table}[h]
\centering
\caption{ERA5 datasets and climate classifications (approximate K\"{o}ppen codes of the nearest city~\cite{beck2018koppen}).}
\label{tab:datasets}
\small
\begin{tabular}{
  >{\raggedright\arraybackslash}p{0.19\linewidth}
  >{\raggedright\arraybackslash}p{0.09\linewidth}
  >{\raggedright\arraybackslash}p{0.24\linewidth}
  >{\raggedright\arraybackslash}p{0.36\linewidth}}
\toprule
Location & K\"{o}ppen & Climate regime & Primary challenge \\
\midrule
Ankara, Turkey  & BSk & Semi-arid continental & Low humidity, moderate variability \\
Dubai, UAE      & BWh & Hot desert            & Extreme aridity, humidity saturation \\
Bergen, Norway  & Cfb & Oceanic               & High precipitation, complex coupling \\
Yakutsk, Russia & Dfc & Subarctic             & Extreme cold, widest seasonal range \\
Lhasa, China    & BSk & High-altitude semi-arid & Altitude effects \\
\bottomrule
\end{tabular}
\end{table}

\noindent
\textbf{Features} ($F=\protoNFeatures$): 2\,m temperature, 2\,m dewpoint, mean
sea-level pressure, wind speed, wind direction (sine and cosine
encoding), relative humidity, hour-of-day (sine and cosine encoding),
vapour pressure deficit, and \codeTendLag-hour pressure tendency.  Sequences of
length $T=\protoSeqLen$ (one week of hourly data) are extracted with
stride \protoStride{}
separately within the training and evaluation blocks.  The \protoNScaledWord{} non-cyclical features are scaled to $[-1,1]$ with a min--max transform
fitted on the training block; the four sine and cosine encodings are
left unscaled.  All distributional metrics are computed on the
\protoNScaledWord{} scaled features in physical units.  The record lengths and split are
identical across locations, giving $N_{\text{train}} = \protoNTrainSeq$ and
$N_{\text{val}} = \protoNValSeq$ sequences per location; the held-out
block spans \protoValYearsWord{} calendar years including one leap year
(\protoNValHours{} hours).

\subsection{Baselines}
\label{sec:baselines}

We compare against re-implementations of \textbf{RCGAN}
\cite{esteban2017rcgan} and \textbf{TimeGAN} \cite{yoon2019timegan},
evaluated under exactly the pipeline used for StatD2GAN: identical
splits and embargo, calibration fitted on the training block, identical
metrics on the held-out block, and the same five seeds.  Raw
(pre-calibration) diagnostics are recorded for both.

\paragraph{RCGAN.}
The unconditional recurrent core of \cite{esteban2017rcgan}: a
\protoRcLayers-layer LSTM generator (hidden \protoRcHidden, step-wise
latent input of dimension \protoRcLatent, $\tanh$ output) against a
\protoRcLayers-layer LSTM discriminator producing per-step logits,
trained with the BCE objective, non-saturating for the generator (Adam,
learning rate $\protoRcLr$, \protoRcEpochs{} epochs).  RCGAN serves as an
under-performing reference on cross-variable structure, against which
the discriminative power of each metric can be checked.

\paragraph{TimeGAN.}
A compact re-implementation of \cite{yoon2019timegan} with its five
components (embedder, recovery, generator, supervisor, latent
discriminator), each a GRU of hidden size \protoTgHidden{} with
\protoTgLayers{} layers (\codeTgSupLayers{} for the supervisor), trained
in the original three phases: autoencoding (\protoTgItersAe{}
iterations), supervised next-step prediction (\protoTgItersSup), and
joint adversarial training (\protoTgItersJoint) with the moment-matching
and supervised losses, and the rule of skipping discriminator updates
when its loss falls below \codeTgDThresh.

\subsection{Ablation Configurations}
\label{sec:ablations}

Nine arms are trained on every location and seed, giving
$\nLocs \times \protoNArms \times \nSeeds = \nRunsAbl$ runs:
\texttt{full\_model};
the component removals \texttt{no\_D\_temporal}, \texttt{no\_D\_stat},
\texttt{no\_D\_sorted}, \texttt{no\_calibration},
\texttt{no\_aux\_losses};
\texttt{fixed\_lambda} (uniform weights, no replicator updates); and
two component substitutions, \texttt{moment\_D\_stat}
(Section~\ref{sec:discriminators}) and \texttt{rank\_D\_sort}
(the mechanism probe of Section~\ref{sec:discriminators}).
Component-removed variants retain all other components unchanged.

\subsection{Evaluation Metrics}
\label{sec:metrics}

All stochastic subsampling inside the metrics uses one fixed seed,
independent of the training seed, so that metric noise does not
contaminate seed-to-seed variance.

\paragraph{KS sup-distance against a noise-and-shift floor $\downarrow$.}
The primary marginal metric is the two-sample Kolmogorov--Smirnov
sup-distance averaged over the seven scaled features:
\begin{equation}
  \mathrm{KS}
    = \frac{1}{\protoNScaled}\sum_{f=1}^{\protoNScaled}
      \sup_x \abs{\hat{F}_f^{\mathrm{syn}}(x) - \hat{F}_f^{\mathrm{val}}(x)}
  \label{eq:ks}
\end{equation}
Because two finite samples from the same distribution have a nonzero
expected sup-distance, we report KS relative to a per-location
\emph{noise-and-shift floor}: the same statistic computed between the real
training and real held-out blocks. This reference conflates two sources
that a generator cannot be expected to beat: finite-sample noise, common
to every location, and train-to-held-out distribution shift, which
varies by location (Limitation~3). It is a practical reference, not a lower bound,
for calibrated marginal comparison, and not a universal noise level: of
the \phKsRatioCells{} calibrated StatD2GAN runs (every ablation arm except
\texttt{no\_calibration}, all seeds), \phKsRatioCellsBelowOne{} lie below
their location's floor.
Floors are \ksFloorDubai{} (Dubai), \ksFloorBergen{} (Bergen) and
\ksFloorAnkara{} (Ankara), with Yakutsk at \ksFloorYakutsk{} and Lhasa at
\ksFloorLhasa{}
(Table~\ref{tab:floor}).  A threshold-based pass rate is not reported.

Proposition~\ref{prop:lock} (Section~\ref{sec:results_floor}) relates the
calibrated KS to the floor through the distance of the calibrated sample to
the training block.

\paragraph{Kendall $\tau$ MAE $\downarrow$.}
\begin{equation}
  \tau\text{-MAE}
    = \frac{1}{\abs{\mathcal{P}}}\sum_{(i,j)\in\mathcal{P}}
      \abs{\hat{\tau}_{ij}^{\mathrm{syn}} - \hat{\tau}_{ij}^{\mathrm{val}}}
  \label{eq:tau_mae}
\end{equation}
over six physically motivated variable pairs
(temperature--dewpoint, temperature--humidity, pressure--VPD,
humidity--VPD, temperature--pressure, dewpoint--humidity), estimated on
a fixed-seed subsample of \protoTauSubsample{} time points.

\paragraph{ACF MAE $\downarrow$.}
The autocorrelation function is computed per sequence, on a fixed
random subset of \protoAcfNSeq{} sequences, and averaged over sequences
before differencing, for lags $0$ to $\protoAcfLags$ and the
\protoNAcfFeaturesWord{} continuous features (temperature, dewpoint, pressure, wind speed,
humidity):
\begin{equation}
  \mathrm{ACF\text{-}MAE}
    = \frac{1}{\protoNAcfFeatures \cdot (\protoAcfLags + 1)}\sum_{f}\sum_{k=0}^{\protoAcfLags}
      \abs{\bar{\rho}_f^{\mathrm{syn}}(k) - \bar{\rho}_f^{\mathrm{val}}(k)}
  \label{eq:acf_mae}
\end{equation}
The lag-0 term is zero, so the metric equals $\protoAcfLags/(\protoAcfLags+1)$
times the mean absolute difference over lags $1$ to $\protoAcfLags$.

\paragraph{Physical violation rate $\downarrow$.}
The fraction of (sequence, time-step) cells violating at least one of
C1 to C3:
\begin{equation}
  \mathrm{PhysViol}
    = \frac{1}{N_{\mathrm{syn}}\, T}
      \sum_{n=1}^{N_{\mathrm{syn}}}\sum_{t=1}^{T}
      \mathbf{1}\!\left[\exists\,j : c_j\!\left(\tilde{\mathbf{x}}^{(n)}\right)_t
      \text{ violated}\right]
  \label{eq:phys_viol}
\end{equation}
evaluated with the same constants as the training penalty
(Section~\ref{sec:physics}).  We report it both before and after
calibration, which is what localises the source of violations in
Section~\ref{sec:results_cal}.

\paragraph{Sequence-level variance decomposition.}
For temperature, each sequence's \protoSeqLen-hour mean is computed, and we
report the standard deviation of these per-sequence means
(\emph{between-sequence spread}, in kelvin), alongside the mean
within-sequence standard deviation and the mean absolute one-step
difference.  These statistics decompose the pooled variance that the metrics above
aggregate over; they are reported descriptively, and rule D6 uses the
between-sequence spread.

\subsection{Statistical Methodology}
\label{sec:stats}

The unit of evidence is the matched pair: for each contrast
(an ablation arm or baseline against \texttt{full\_model}), differences
are formed within each (location, seed) cell, giving $n=\nPairs$ pairs per contrast.  Each contrast is tested with the two-sided Wilcoxon
signed-rank test \cite{wilcoxon1945}, and within each metric the
resulting $p$-values are Holm-corrected \cite{holm1979} across the
\protoNContrastsWord{} contrasts; the \protoNBaselinesWord{} baseline
contrasts form a separate Holm family.  We report median paired differences alongside
corrected $p$-values, and we report null results with the same
prominence as significant ones.

Run-to-run training noise sets the resolution limit of any comparison.
Between identically configured runs differing only in seed, the median
absolute $\tau$-MAE difference is \runNoiseDtau.  Single-seed
comparisons at or below this scale are meaningless, and none are made
in this paper.

\begin{figure}[t]
\centering
\includegraphics[width=0.85\linewidth]{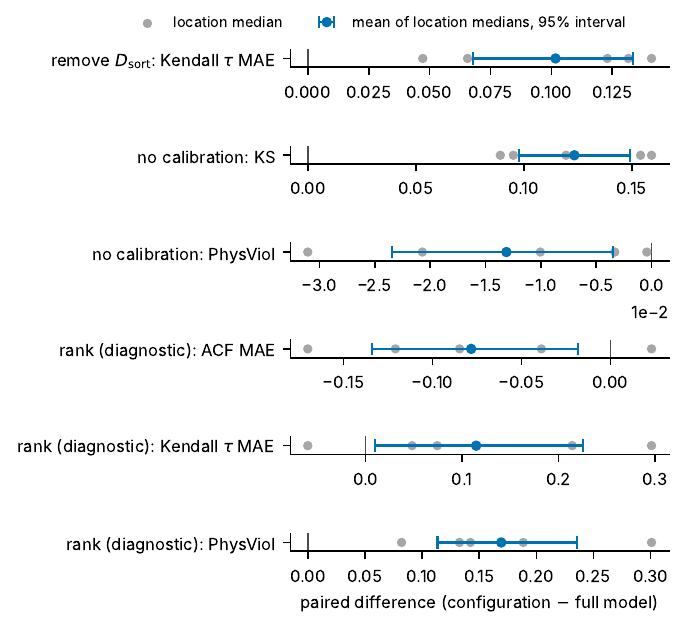}
\caption{Location-clustered robustness check for the \nHolmSigWord{}
Holm-significant contrasts behind the paper's headline claims. Points:
mean of the \nLocsProtoWord{} per-location median differences
(configuration $-$ \texttt{full\_model}); grey dots: the per-location
medians.
Whiskers: \protoBootCI\% bootstrap interval from resampling the
\nLocsProtoWord{} location-level medians with replacement (\protoNBoot{}
draws). All \nHolmSigWord{} intervals exclude zero.}
\label{fig:location_robustness}
\end{figure}

The \nPairs{} matched pairs are not fully independent: the
\protoNSeedsWord{} seeds within a location share the same
training/held-out realisation of that location's ERA5 record, so the design has \nLocsProtoWord{} independent geographic units, not \nPairs.
The primary inference is therefore conditional on this fixed
\nLocsProtoWord{}-location benchmark: the Wilcoxon and Holm results describe the
locations studied, and the location-clustered analysis below assesses
sensitivity to dependence among seeds within a location, not generalisation
to unseen climates. To check that the Wilcoxon result is not an
artefact of treating correlated seeds as independent, we collapse
each contrast to one median difference per location and resample the
\nLocsProtoWord{} location-level medians with replacement (\protoNBoot{}
draws). Every contrast that survives Holm correction in
Table~\ref{tab:wilcoxon} also has a location-clustered
\protoBootCI\% bootstrap interval that excludes zero
(Fig.~\ref{fig:location_robustness}; e.g.\ \texttt{no\_calibration} on
KS, $[\ciLoNoCalibrationKs, \ciHiNoCalibrationKs]$;
\texttt{rank\_D\_sort} on $\tau$-MAE,
$[\ciLoRankDSortTau, \ciHiRankDSortTau]$; \texttt{no\_D\_sorted} on
$\tau$-MAE, $[\ciLoNoDSortedTau, \ciHiNoDSortedTau]$), descriptively consistent with
the direction and approximate magnitude of each effect under the
five-unit view.  With \nLocs{} clusters the bootstrap has at most
\nBootDistinct{} distinct resamples, so these intervals are coarse. An exact sign test across five locations
cannot itself reach conventional significance even at full agreement ($p=2^{1-\nLocs}=\protoSignTestPMin$ two-sided); this is a ceiling of
$n=\nLocs$, not a failure of the data, and we report it for transparency rather than relying on
it.

\subsection{Hyperparameters}

Table~\ref{tab:hyperparams} lists all model and training
hyperparameters, matched one-to-one to the released training code.

\begin{table}[h]
\centering
\caption{Model and training hyperparameters.}
\label{tab:hyperparams}
\begin{tabular}{lll}
\toprule
Category & Parameter & Value \\
\midrule
\multirow{4}{*}{Data}
  & Sequence length $T$          & \protoSeqLen{} (hourly, stride \protoStride) \\
  & Features $F$                 & \protoNFeatures{} (\protoNScaled{} scaled to $[-1,1]$) \\
  & Split                        & last \protoValYears{} calendar years held out, \protoEmbargoHours\,h embargo \\
  & Noise dimension $d_z$        & \protoLatent{} \\
\midrule
\multirow{4}{*}{Model}
  & Hidden dimension             & \protoHidden{} \\
  & $\lambda_k^{(0)}$            & $1$ \\
  & $[\lambda_{\min},\lambda_{\max}]$, ratio cap & $[\protoEvoLo,\ \protoEvoHi]$, $\protoEvoRatio$ \\
  & $\lambda_{\mathrm{GP}}$ (lazy, every \protoGpFreq{}th batch) & \protoGpLambda{} (effective) \\
\midrule
\multirow{6}{*}{Training}
  & Optimiser                    & Adam ($\beta_1{=}\codeAdamBetaOne$, $\beta_2{=}\codeAdamBetaTwo$) \\
  & Learning rate (G\,=\,D)      & $\protoLrG$ (G divided by \codeLrDecayDiv{} at epoch \codeLrDecayEp) \\
  & Batch size                   & \protoBatch{} \\
  & Epochs (of which warmup)     & \protoEpochs{} (\protoWarmup) \\
  & Critic steps $n_{\mathrm{critic}}$ & \codeNCritic{} (\codeNCriticLate{} after epoch \codeNCriticEp) \\
  & Replicator $\eta$            & \protoEvoEta{} \\
\midrule
\multirow{2}{*}{Evaluation}
  & Generated sample size        & \protoNEval{} sequences \\
  & Metric subsamples            & $\tau$: \protoTauSubsample; fixed metric seed \\
\bottomrule
\end{tabular}
\end{table}

\section{Results}
\label{sec:results}

\begin{table}[t]
\centering\small
\caption{KS sup-distance relative to the real train versus held-out reference (mean over 5 seeds). The reference is the mean over the 7 scalable features of the KS distance between real training and real held-out marginals. It is a practical reference, not a mathematical lower bound.}
\label{tab:floor}
\begin{tabular}{lrrr}
\toprule
Location & Reference & Full model (ratio) & No calibration (ratio) \\
\midrule
Ankara & 0.0363 & 1.02 & 5.85 \\
Bergen & 0.0360 & 1.03 & 4.85 \\
Dubai & 0.0344 & 1.02 & 5.44 \\
Lhasa & 0.0846 & 1.00 & 2.12 \\
Yakutsk & 0.0488 & 1.04 & 3.48 \\
\bottomrule
\end{tabular}
\end{table}

\begin{table}[t]
\centering\small
\caption{KS sup-distance (mean $\pm$ SD over 5 seeds, held-out).}
\label{tab:ks_stat}
\begin{tabular}{lrrrrr}
\toprule
Arm & Ankara & Bergen & Dubai & Lhasa & Yakutsk \\
\midrule
fixed\_lambda & 0.038 $\pm$ 0.002 & 0.037 $\pm$ 0.001 & 0.036 $\pm$ 0.001 & 0.085 $\pm$ 0.002 & 0.052 $\pm$ 0.002 \\
full\_model & 0.037 $\pm$ 0.001 & 0.037 $\pm$ 0.000 & 0.035 $\pm$ 0.001 & 0.085 $\pm$ 0.001 & 0.051 $\pm$ 0.002 \\
moment\_D\_stat & 0.038 $\pm$ 0.001 & 0.038 $\pm$ 0.001 & 0.035 $\pm$ 0.001 & 0.084 $\pm$ 0.001 & 0.053 $\pm$ 0.004 \\
no\_D\_sorted & 0.039 $\pm$ 0.003 & 0.037 $\pm$ 0.001 & 0.037 $\pm$ 0.003 & 0.086 $\pm$ 0.003 & 0.049 $\pm$ 0.004 \\
no\_D\_stat & 0.038 $\pm$ 0.001 & 0.037 $\pm$ 0.001 & 0.036 $\pm$ 0.001 & 0.085 $\pm$ 0.001 & 0.048 $\pm$ 0.002 \\
no\_D\_temporal & 0.036 $\pm$ 0.001 & 0.037 $\pm$ 0.001 & 0.036 $\pm$ 0.001 & 0.084 $\pm$ 0.002 & 0.053 $\pm$ 0.005 \\
no\_aux\_losses & 0.038 $\pm$ 0.001 & 0.037 $\pm$ 0.001 & 0.036 $\pm$ 0.001 & 0.085 $\pm$ 0.001 & 0.050 $\pm$ 0.001 \\
no\_calibration & 0.212 $\pm$ 0.060 & 0.175 $\pm$ 0.038 & 0.187 $\pm$ 0.081 & 0.180 $\pm$ 0.036 & 0.170 $\pm$ 0.054 \\
rank\_D\_sort & 0.038 $\pm$ 0.001 & 0.036 $\pm$ 0.001 & 0.035 $\pm$ 0.000 & 0.085 $\pm$ 0.000 & 0.051 $\pm$ 0.001 \\
\bottomrule
\end{tabular}
\end{table}

\begin{table}[t]
\centering\small
\caption{Kendall tau MAE (mean $\pm$ SD over 5 seeds, held-out).}
\label{tab:kendall_mae}
\begin{tabular}{lrrrrr}
\toprule
Arm & Ankara & Bergen & Dubai & Lhasa & Yakutsk \\
\midrule
fixed\_lambda & 0.163 $\pm$ 0.044 & 0.238 $\pm$ 0.060 & 0.158 $\pm$ 0.034 & 0.153 $\pm$ 0.055 & 0.122 $\pm$ 0.056 \\
full\_model & 0.200 $\pm$ 0.023 & 0.214 $\pm$ 0.032 & 0.161 $\pm$ 0.085 & 0.203 $\pm$ 0.054 & 0.143 $\pm$ 0.080 \\
moment\_D\_stat & 0.163 $\pm$ 0.051 & 0.180 $\pm$ 0.056 & 0.152 $\pm$ 0.056 & 0.118 $\pm$ 0.041 & 0.119 $\pm$ 0.044 \\
no\_D\_sorted & 0.279 $\pm$ 0.116 & 0.249 $\pm$ 0.084 & 0.315 $\pm$ 0.081 & 0.229 $\pm$ 0.066 & 0.260 $\pm$ 0.025 \\
no\_D\_stat & 0.242 $\pm$ 0.106 & 0.227 $\pm$ 0.083 & 0.179 $\pm$ 0.034 & 0.179 $\pm$ 0.030 & 0.148 $\pm$ 0.065 \\
no\_D\_temporal & 0.250 $\pm$ 0.124 & 0.193 $\pm$ 0.063 & 0.224 $\pm$ 0.087 & 0.156 $\pm$ 0.073 & 0.145 $\pm$ 0.045 \\
no\_aux\_losses & 0.272 $\pm$ 0.151 & 0.242 $\pm$ 0.102 & 0.188 $\pm$ 0.108 & 0.180 $\pm$ 0.041 & 0.144 $\pm$ 0.016 \\
no\_calibration & 0.200 $\pm$ 0.023 & 0.214 $\pm$ 0.032 & 0.161 $\pm$ 0.085 & 0.203 $\pm$ 0.054 & 0.143 $\pm$ 0.080 \\
rank\_D\_sort & 0.285 $\pm$ 0.063 & 0.207 $\pm$ 0.090 & 0.324 $\pm$ 0.029 & 0.231 $\pm$ 0.048 & 0.419 $\pm$ 0.107 \\
\bottomrule
\end{tabular}
\end{table}

\begin{table}[t]
\centering\small
\caption{ACF MAE (mean $\pm$ SD over 5 seeds, held-out).}
\label{tab:acf_mae}
\begin{tabular}{lrrrrr}
\toprule
Arm & Ankara & Bergen & Dubai & Lhasa & Yakutsk \\
\midrule
fixed\_lambda & 0.317 $\pm$ 0.033 & 0.319 $\pm$ 0.125 & 0.284 $\pm$ 0.090 & 0.205 $\pm$ 0.024 & 0.361 $\pm$ 0.089 \\
full\_model & 0.276 $\pm$ 0.066 & 0.431 $\pm$ 0.025 & 0.286 $\pm$ 0.053 & 0.215 $\pm$ 0.086 & 0.378 $\pm$ 0.020 \\
moment\_D\_stat & 0.308 $\pm$ 0.039 & 0.375 $\pm$ 0.132 & 0.307 $\pm$ 0.063 & 0.228 $\pm$ 0.090 & 0.410 $\pm$ 0.053 \\
no\_D\_sorted & 0.273 $\pm$ 0.055 & 0.416 $\pm$ 0.086 & 0.245 $\pm$ 0.043 & 0.197 $\pm$ 0.056 & 0.411 $\pm$ 0.020 \\
no\_D\_stat & 0.320 $\pm$ 0.038 & 0.392 $\pm$ 0.130 & 0.273 $\pm$ 0.054 & 0.232 $\pm$ 0.067 & 0.275 $\pm$ 0.129 \\
no\_D\_temporal & 0.274 $\pm$ 0.032 & 0.390 $\pm$ 0.024 & 0.268 $\pm$ 0.079 & 0.260 $\pm$ 0.046 & 0.385 $\pm$ 0.020 \\
no\_aux\_losses & 0.263 $\pm$ 0.037 & 0.337 $\pm$ 0.098 & 0.256 $\pm$ 0.056 & 0.257 $\pm$ 0.069 & 0.280 $\pm$ 0.100 \\
no\_calibration & 0.273 $\pm$ 0.068 & 0.435 $\pm$ 0.036 & 0.274 $\pm$ 0.049 & 0.210 $\pm$ 0.085 & 0.368 $\pm$ 0.017 \\
rank\_D\_sort & 0.272 $\pm$ 0.016 & 0.263 $\pm$ 0.064 & 0.232 $\pm$ 0.052 & 0.219 $\pm$ 0.067 & 0.263 $\pm$ 0.086 \\
\bottomrule
\end{tabular}
\end{table}

\begin{table}[t]
\centering\small
\caption{Physics violation rate (mean $\pm$ SD over 5 seeds, held-out).}
\label{tab:phys_viol_rate}
\begin{tabular}{lrrrrr}
\toprule
Arm & Ankara & Bergen & Dubai & Lhasa & Yakutsk \\
\midrule
fixed\_lambda & 0.043 $\pm$ 0.058 & 0.081 $\pm$ 0.062 & 0.005 $\pm$ 0.012 & 0.017 $\pm$ 0.029 & 0.018 $\pm$ 0.018 \\
full\_model & 0.024 $\pm$ 0.027 & 0.063 $\pm$ 0.075 & 0.035 $\pm$ 0.061 & 0.056 $\pm$ 0.120 & 0.078 $\pm$ 0.092 \\
moment\_D\_stat & 0.020 $\pm$ 0.022 & 0.068 $\pm$ 0.063 & 0.000 $\pm$ 0.001 & 0.012 $\pm$ 0.019 & 0.017 $\pm$ 0.018 \\
no\_D\_sorted & 0.133 $\pm$ 0.145 & 0.091 $\pm$ 0.091 & 0.044 $\pm$ 0.040 & 0.037 $\pm$ 0.047 & 0.171 $\pm$ 0.101 \\
no\_D\_stat & 0.089 $\pm$ 0.123 & 0.020 $\pm$ 0.004 & 0.001 $\pm$ 0.001 & 0.025 $\pm$ 0.045 & 0.040 $\pm$ 0.048 \\
no\_D\_temporal & 0.023 $\pm$ 0.038 & 0.021 $\pm$ 0.007 & 0.010 $\pm$ 0.023 & 0.010 $\pm$ 0.016 & 0.022 $\pm$ 0.019 \\
no\_aux\_losses & 0.021 $\pm$ 0.013 & 0.087 $\pm$ 0.085 & 0.027 $\pm$ 0.032 & 0.033 $\pm$ 0.043 & 0.065 $\pm$ 0.073 \\
no\_calibration & 0.005 $\pm$ 0.006 & 0.006 $\pm$ 0.008 & 0.000 $\pm$ 0.000 & 0.000 $\pm$ 0.000 & 0.000 $\pm$ 0.000 \\
rank\_D\_sort & 0.115 $\pm$ 0.098 & 0.254 $\pm$ 0.076 & 0.099 $\pm$ 0.028 & 0.196 $\pm$ 0.069 & 0.356 $\pm$ 0.139 \\
\bottomrule
\end{tabular}
\end{table}

\begin{table}[t]
\centering\small
\caption{Paired Wilcoxon signed-rank over 25 (location, seed) pairs, two-sided; Holm step-down adjustment (monotone) within each metric family. Difference = arm minus full model on the same seed.}
\label{tab:wilcoxon}
\begin{tabular}{llrrrl}
\toprule
Metric & Contrast & Median diff & p & p (Holm) & Positive pairs by location \\
\midrule
ks\_stat & fixed\_lambda & -0.0001 & 0.578 & 1.000 & AN:3/5 BE:1/5 DU:3/5 LH:2/5 YA:3/5 \\
ks\_stat & moment\_D\_stat & 0.0003 & 0.241 & 1.000 & AN:5/5 BE:2/5 DU:2/5 LH:3/5 YA:3/5 \\
ks\_stat & no\_D\_sorted & -0.0001 & 0.578 & 1.000 & AN:4/5 BE:1/5 DU:3/5 LH:3/5 YA:1/5 \\
ks\_stat & no\_D\_stat & 0.0004 & 0.634 & 1.000 & AN:4/5 BE:3/5 DU:4/5 LH:3/5 YA:1/5 \\
ks\_stat & no\_D\_temporal & 0.0001 & 1.000 & 1.000 & AN:2/5 BE:2/5 DU:3/5 LH:2/5 YA:4/5 \\
ks\_stat & no\_aux\_losses & -0.0000 & 0.791 & 1.000 & AN:5/5 BE:1/5 DU:3/5 LH:3/5 YA:0/5 \\
ks\_stat & no\_calibration & 0.1259 & <0.001 & <0.001 & AN:5/5 BE:5/5 DU:5/5 LH:5/5 YA:5/5 \\
ks\_stat & rank\_D\_sort & 0.0002 & 0.525 & 1.000 & AN:4/5 BE:1/5 DU:2/5 LH:3/5 YA:3/5 \\
kendall\_mae & fixed\_lambda & -0.0019 & 0.474 & 1.000 & AN:1/5 BE:4/5 DU:3/5 LH:1/5 YA:3/5 \\
kendall\_mae & moment\_D\_stat & -0.0220 & 0.034 & 0.205 & AN:1/5 BE:1/5 DU:2/5 LH:2/5 YA:3/5 \\
kendall\_mae & no\_D\_sorted & 0.1155 & <0.001 & 0.002 & AN:4/5 BE:3/5 DU:5/5 LH:3/5 YA:5/5 \\
kendall\_mae & no\_D\_stat & 0.0319 & 0.751 & 1.000 & AN:3/5 BE:3/5 DU:4/5 LH:1/5 YA:3/5 \\
kendall\_mae & no\_D\_temporal & -0.0072 & 0.791 & 1.000 & AN:3/5 BE:2/5 DU:4/5 LH:0/5 YA:3/5 \\
kendall\_mae & no\_aux\_losses & 0.0160 & 0.426 & 1.000 & AN:3/5 BE:4/5 DU:4/5 LH:3/5 YA:3/5 \\
kendall\_mae & no\_calibration & 0.0000 & 0.826 & 1.000 & AN:0/5 BE:1/5 DU:3/5 LH:2/5 YA:2/5 \\
kendall\_mae & rank\_D\_sort & 0.0943 & <0.001 & 0.006 & AN:5/5 BE:1/5 DU:4/5 LH:3/5 YA:5/5 \\
acf\_mae & fixed\_lambda & 0.0010 & 0.653 & 1.000 & AN:3/5 BE:2/5 DU:2/5 LH:3/5 YA:3/5 \\
acf\_mae & moment\_D\_stat & 0.0069 & 0.634 & 1.000 & AN:3/5 BE:1/5 DU:4/5 LH:2/5 YA:4/5 \\
acf\_mae & no\_D\_sorted & 0.0125 & 0.979 & 1.000 & AN:3/5 BE:3/5 DU:2/5 LH:4/5 YA:4/5 \\
acf\_mae & no\_D\_stat & -0.0140 & 0.426 & 1.000 & AN:2/5 BE:3/5 DU:2/5 LH:2/5 YA:1/5 \\
acf\_mae & no\_D\_temporal & -0.0030 & 0.916 & 1.000 & AN:2/5 BE:0/5 DU:3/5 LH:4/5 YA:3/5 \\
acf\_mae & no\_aux\_losses & -0.0475 & 0.067 & 0.400 & AN:3/5 BE:1/5 DU:2/5 LH:4/5 YA:1/5 \\
acf\_mae & no\_calibration & -0.0033 & 0.007 & 0.052 & AN:1/5 BE:2/5 DU:2/5 LH:1/5 YA:0/5 \\
acf\_mae & rank\_D\_sort & -0.0848 & 0.003 & 0.025 & AN:2/5 BE:0/5 DU:1/5 LH:3/5 YA:1/5 \\
phys\_viol\_rate & fixed\_lambda & -0.0001 & 0.426 & 0.852 & AN:3/5 BE:3/5 DU:0/5 LH:2/5 YA:2/5 \\
phys\_viol\_rate & moment\_D\_stat & -0.0003 & 0.156 & 0.625 & AN:3/5 BE:2/5 DU:0/5 LH:1/5 YA:2/5 \\
phys\_viol\_rate & no\_D\_sorted & 0.0406 & 0.059 & 0.294 & AN:3/5 BE:2/5 DU:3/5 LH:3/5 YA:5/5 \\
phys\_viol\_rate & no\_D\_stat & -0.0029 & 0.048 & 0.290 & AN:3/5 BE:0/5 DU:0/5 LH:1/5 YA:2/5 \\
phys\_viol\_rate & no\_D\_temporal & -0.0014 & 0.156 & 0.625 & AN:2/5 BE:0/5 DU:2/5 LH:3/5 YA:3/5 \\
phys\_viol\_rate & no\_aux\_losses & 0.0006 & 0.874 & 0.874 & AN:3/5 BE:3/5 DU:2/5 LH:3/5 YA:3/5 \\
phys\_viol\_rate & no\_calibration & -0.0100 & <0.001 & <0.001 & AN:2/5 BE:1/5 DU:0/5 LH:0/5 YA:0/5 \\
phys\_viol\_rate & rank\_D\_sort & 0.1423 & <0.001 & <0.001 & AN:4/5 BE:5/5 DU:4/5 LH:4/5 YA:5/5 \\
\bottomrule
\end{tabular}
\end{table}

\begin{table}[t]
\centering\small
\caption{Paired Wilcoxon signed-rank over 25 (location, seed) pairs, two-sided; Holm step-down adjustment (monotone) within each metric family. Baselines against the full model, family of two.}
\label{tab:baselines}
\begin{tabular}{llrrrl}
\toprule
Metric & Contrast & Median diff & p & p (Holm) & Positive pairs by location \\
\midrule
ks\_stat & rcgan & -0.0008 & 0.210 & 0.420 & AN:1/5 BE:1/5 DU:3/5 LH:3/5 YA:2/5 \\
ks\_stat & timegan & 0.0001 & 0.634 & 0.634 & AN:4/5 BE:3/5 DU:2/5 LH:2/5 YA:2/5 \\
kendall\_mae & rcgan & 0.1879 & <0.001 & <0.001 & AN:5/5 BE:4/5 DU:5/5 LH:3/5 YA:5/5 \\
kendall\_mae & timegan & -0.0771 & 0.002 & 0.002 & AN:0/5 BE:1/5 DU:0/5 LH:0/5 YA:3/5 \\
acf\_mae & rcgan & -0.0068 & 0.833 & 0.833 & AN:4/5 BE:1/5 DU:2/5 LH:2/5 YA:2/5 \\
acf\_mae & timegan & -0.1690 & <0.001 & <0.001 & AN:1/5 BE:0/5 DU:0/5 LH:0/5 YA:1/5 \\
phys\_viol\_rate & rcgan & 0.1216 & <0.001 & <0.001 & AN:4/5 BE:5/5 DU:4/5 LH:4/5 YA:4/5 \\
phys\_viol\_rate & timegan & 0.0063 & 0.596 & 0.596 & AN:4/5 BE:3/5 DU:3/5 LH:4/5 YA:3/5 \\
\bottomrule
\end{tabular}
\end{table}

\begin{table}[t]
\centering\small
\caption{Location-clustered bootstrap of the paired difference. Statistic: mean over the 5 locations of the per-location median of 5 paired seed differences; locations resampled with replacement, 20000 draws, percentile 95\% interval. With five clusters the interval is coarse.}
\label{tab:bootstrap}
\begin{tabular}{lrrrr}
\toprule
Metric & Contrast & Point & Lo & Hi \\
\midrule
ks\_stat & fixed\_lambda & 0.0005 & -0.0001 & 0.0010 \\
ks\_stat & moment\_D\_stat & 0.0006 & -0.0002 & 0.0016 \\
ks\_stat & no\_D\_sorted & 0.0000 & -0.0012 & 0.0011 \\
ks\_stat & no\_D\_stat & -0.0000 & -0.0015 & 0.0009 \\
ks\_stat & no\_D\_temporal & -0.0002 & -0.0009 & 0.0006 \\
ks\_stat & no\_aux\_losses & 0.0001 & -0.0002 & 0.0005 \\
ks\_stat & no\_calibration & 0.1234 & 0.0977 & 0.1492 \\
ks\_stat & rank\_D\_sort & 0.0005 & -0.0001 & 0.0012 \\
kendall\_mae & fixed\_lambda & 0.0057 & -0.0245 & 0.0334 \\
kendall\_mae & moment\_D\_stat & -0.0360 & -0.0802 & -0.0033 \\
kendall\_mae & no\_D\_sorted & 0.1017 & 0.0678 & 0.1337 \\
kendall\_mae & no\_D\_stat & 0.0263 & -0.0068 & 0.0490 \\
kendall\_mae & no\_D\_temporal & 0.0250 & -0.0173 & 0.0726 \\
kendall\_mae & no\_aux\_losses & 0.0237 & 0.0155 & 0.0319 \\
kendall\_mae & no\_calibration & 0.0000 & 0.0000 & 0.0000 \\
kendall\_mae & rank\_D\_sort & 0.1146 & 0.0101 & 0.2251 \\
acf\_mae & fixed\_lambda & -0.0149 & -0.0505 & 0.0083 \\
acf\_mae & moment\_D\_stat & 0.0021 & -0.0165 & 0.0200 \\
acf\_mae & no\_D\_sorted & 0.0109 & -0.0214 & 0.0328 \\
acf\_mae & no\_D\_stat & -0.0448 & -0.1135 & -0.0015 \\
acf\_mae & no\_D\_temporal & 0.0116 & -0.0162 & 0.0483 \\
acf\_mae & no\_aux\_losses & -0.0377 & -0.0964 & 0.0119 \\
acf\_mae & no\_calibration & -0.0052 & -0.0072 & -0.0033 \\
acf\_mae & rank\_D\_sort & -0.0783 & -0.1340 & -0.0181 \\
phys\_viol\_rate & fixed\_lambda & -0.0012 & -0.0039 & 0.0013 \\
phys\_viol\_rate & moment\_D\_stat & -0.0028 & -0.0074 & 0.0005 \\
phys\_viol\_rate & no\_D\_sorted & 0.0278 & -0.0041 & 0.0596 \\
phys\_viol\_rate & no\_D\_stat & -0.0047 & -0.0090 & -0.0007 \\
phys\_viol\_rate & no\_D\_temporal & -0.0058 & -0.0121 & -0.0004 \\
phys\_viol\_rate & no\_aux\_losses & 0.0022 & -0.0009 & 0.0066 \\
phys\_viol\_rate & no\_calibration & -0.0131 & -0.0234 & -0.0035 \\
phys\_viol\_rate & rank\_D\_sort & 0.1693 & 0.1135 & 0.2355 \\
\bottomrule
\end{tabular}
\end{table}

\begin{table}[t]
\centering\small
\caption{Physics violation rate and KS sup-distance for the full model: raw generator output, after calibration, after projection (mean over seeds).}
\label{tab:projection}
\begin{tabular}{lrrrrr}
\toprule
Location & Viol. raw & Viol. calibrated & Viol. projected & KS calibrated & KS projected \\
\midrule
Ankara & 0.0048 & 0.0243 & 0.0000 & 0.0369 & 0.0370 \\
Bergen & 0.0056 & 0.0627 & 0.0000 & 0.0372 & 0.0381 \\
Dubai & 0.0001 & 0.0352 & 0.0000 & 0.0352 & 0.0360 \\
Lhasa & 0.0001 & 0.0559 & 0.0000 & 0.0846 & 0.0868 \\
Yakutsk & 0.0002 & 0.0778 & 0.0000 & 0.0507 & 0.0510 \\
\bottomrule
\end{tabular}
\end{table}

\begin{table}[t]
\centering\small
\caption{Sequence-level temperature statistics (K), median over seeds; real = held-out sequences. seq\_mean\_sd: SD of per-sequence means; within\_sd: mean within-sequence SD.}
\label{tab:seq}
\begin{tabular}{lrrrrrrrr}
\toprule
Location & \multicolumn{4}{c}{seq\_mean\_sd} & \multicolumn{4}{c}{within\_sd} \\
\cmidrule(lr){2-5}\cmidrule(lr){6-9}
 & real & StatD2GAN & TimeGAN & RCGAN & real & StatD2GAN & TimeGAN & RCGAN \\
\midrule
Ankara & 8.21 & 7.71 & 8.35 & 0.95 & 4.44 & 5.71 & 4.93 & 9.65 \\
Bergen & 5.18 & 4.22 & 4.43 & 0.59 & 2.14 & 3.58 & 3.25 & 5.50 \\
Dubai & 5.39 & 4.64 & 5.37 & 2.57 & 2.65 & 4.06 & 3.05 & 5.64 \\
Lhasa & 8.65 & 7.42 & 7.75 & 4.52 & 4.53 & 5.22 & 4.71 & 7.94 \\
Yakutsk & 21.41 & 21.38 & 19.45 & 12.99 & 4.78 & 5.88 & 9.13 & 17.06 \\
\bottomrule
\end{tabular}
\end{table}

\begin{table}[t]
\centering\small
\caption{Pre-declared pilot gate (Dubai) and its outcome.}
\label{tab:pilot}
\begin{tabular}{lrrl}
\toprule
Gate & Change in $\tau$ MAE (\%) & Positive seeds & Outcome \\
\midrule
Remove $\Dsort$ (Dubai) & 96.2 & 5/5 & passed \\
\bottomrule
\end{tabular}
\par\smallskip
\begin{minipage}{0.9\linewidth}\footnotesize
Rule: $\overline{\tau\text{-}\mathrm{MAE}}(\texttt{no\_D\_sorted})\,/\,\overline{\tau\text{-}\mathrm{MAE}}(\texttt{full\_model}) - 1 \ge 0.20$, and the paired per-seed difference \texttt{no\_D\_sorted} $-$ \texttt{full\_model} is positive in at least 4 of 5 seeds; means over the 5 pilot seeds at Dubai.
\end{minipage}
\end{table}

\subsection{Calibration Locks the Marginal Metric to the Noise Floor}
\label{sec:results_floor}

\begin{figure}[t]
\centering
\includegraphics[width=0.85\linewidth]{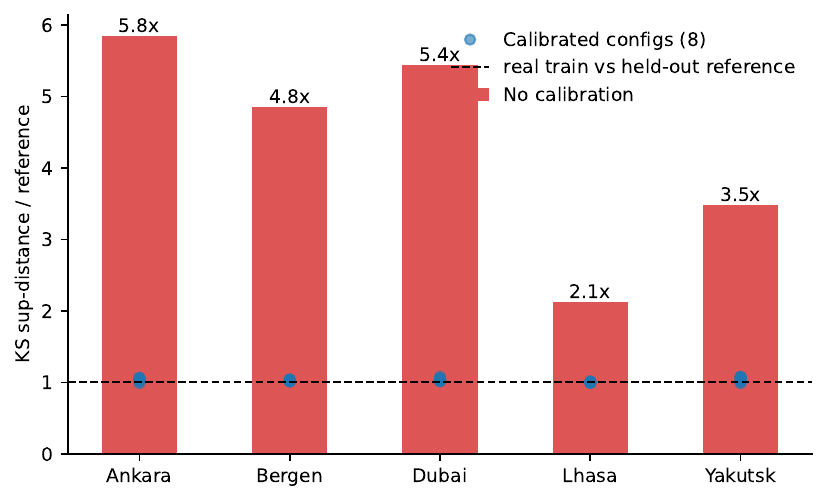}
\caption{Calibration locks the marginal fit to the real train-versus-held-out
noise-and-shift floor.  Bars: \texttt{no\_calibration} KS sup-distance as a
multiple of the per-location floor (mean over seeds).  Dots: the eight
calibrated configurations, which lie close to the dashed reference line
at this scale.}
\label{fig:ks_floor}
\end{figure}

Table~\ref{tab:ks_stat} reports the KS sup-distance for all nine arms,
and Table~\ref{tab:floor} the per-location real-versus-real noise-and-shift floors
together with the full-model and \texttt{no\_calibration} ratios.
The pattern is uniform (Fig.~\ref{fig:ks_floor}): every calibrated
configuration, regardless of which discriminators are present, lands
close to its location's floor (Table~\ref{tab:ks_stat}).  The full model
sits at a median of \ksRatioFullMedian{} times the floor (location means
\ksRatioFullMin{} to \ksRatioFullMax), and \phKsRatioCellsBelowOne{} of the
\phKsRatioCells{} calibrated StatD2GAN runs lie below it.  Removing calibration multiplies the distance
by a factor of \ksRatioNoCalLhasa{} (Lhasa) to \ksRatioNoCalAnkara{} (Ankara) over the floor.  Among the
\protoNContrastsWord{} paired contrasts against the full model, \texttt{no\_calibration}
is the only one significant on KS after Holm correction (median paired
difference $+\medNoCalibrationKs$, $p_{\mathrm{Holm}}\pHolmNoCalibrationKs$; Table~\ref{tab:wilcoxon}); every
architectural contrast has a median KS difference below $10^{-3}$.

The decisive demonstration comes from the baselines.  Neither
re-implemented baseline is distinguishable from StatD2GAN on calibrated
KS ($p_{\mathrm{Holm}}\pHolmBaseRcganKs$ for RCGAN and
$p_{\mathrm{Holm}}\pHolmBaseTimeganKs$ for TimeGAN;
Table~\ref{tab:baselines}), although Section~\ref{sec:results_baselines}
shows that both differ significantly from it on dependence metrics.
A marginal metric computed after calibration therefore certifies the
calibration map, not the generator, and does not separate architectures
at this resolution.  This holds under the held-out protocol, with calibration fitted on the
training block only; it is a property of what isotonic quantile mapping
optimises, not of circular measurement.  All architectural conclusions
below rest on the dependence, autocorrelation, and physics metrics.

\begin{proposition}\label{prop:lock}
Let $d_f(F,G)=\sup_x\lvert F(x)-G(x)\rvert$, and let $\hat F_f^{\mathrm{cal}}$,
$\hat F_f^{\mathrm{train}}$ and $\hat F_f^{\mathrm{val}}$ be the empirical CDFs
of feature $f$ for the calibrated synthetic sample (the evaluation draw after
calibration), the training block and the
held-out block. With $\mathrm{KS}^{\mathrm{cal}}$ the mean over the \protoNScaledWord{} scaled
features of $d_f(\hat F_f^{\mathrm{cal}},\hat F_f^{\mathrm{val}})$, the floor
$\mathrm{KS}^{\mathrm{floor}}$ the same mean of
$d_f(\hat F_f^{\mathrm{train}},\hat F_f^{\mathrm{val}})$, and
$\bar d^{\,\mathrm{cal,train}}$ the mean of
$d_f(\hat F_f^{\mathrm{cal}},\hat F_f^{\mathrm{train}})$,
\[
  \bigl\lvert \mathrm{KS}^{\mathrm{cal}}-\mathrm{KS}^{\mathrm{floor}}\bigr\rvert
  \;\le\; \bar d^{\,\mathrm{cal,train}} .
\]
\end{proposition}
\begin{proof}
$d_f$ is a metric, so $\lvert d_f(\hat F^{\mathrm{cal}},\hat F^{\mathrm{val}})
- d_f(\hat F^{\mathrm{train}},\hat F^{\mathrm{val}})\rvert \le
d_f(\hat F^{\mathrm{cal}},\hat F^{\mathrm{train}})$ for every $f$; average over
$f$ and use $\lvert \operatorname{mean}_f a_f\rvert\le\operatorname{mean}_f\lvert a_f\rvert$.
\end{proof}
The bound holds for any generator, and the architecture enters it only through
$\bar d^{\,\mathrm{cal,train}}$; it gives
$\mathrm{KS}^{\mathrm{cal}}\ge\mathrm{KS}^{\mathrm{floor}}-\bar d^{\,\mathrm{cal,train}}$,
so the floor is a practical reference, not a lower bound.  Because the
calibration map is fitted to the training block on which
$\bar d^{\,\mathrm{cal,train}}$ is computed, the smallness of
$\bar d^{\,\mathrm{cal,train}}$ is in part a consequence of the calibration
target; the proposition does not account for it.  Because both quantile
sequences in Eq.~\eqref{eq:isotonic} are non-decreasing, the minimiser
interpolates them exactly whenever the synthetic quantiles are distinct, so
$\phi_f$ is the piecewise-linear empirical quantile map from the calibration
draw to the training block.  The residual $\bar d^{\,\mathrm{cal,train}}$
therefore reflects the grid of \protoNQuantiles{} levels and the difference
between the independent calibration and evaluation draws rather than an
imperfect fit; it varies across generators (Table~\ref{tab:dist}), so the
bound, not the residual, is generator-free.  Recomputed post-freeze for the \phDistNRuns{}
calibrated runs (\ref{app:distance}, Table~\ref{tab:dist}), the bound
confines $\mathrm{KS}^{\mathrm{cal}}$ to within \phDistRatioMed{} times the floor
at the median (\phDistRatioPninety{} at the 90th percentile, at most
\phDistTopOneRatio{}); the realised deviation is \phDistGapMed{} times the floor
and \phDistGapOverDMed{} times the bound.  In the \phDistBelowAll{} below-floor
runs, which include the \phKsRatioCellsBelowOne{} StatD2GAN runs above, the
shortfall uses at most \phDistBelowMaxRatio{} of the margin
$\bar d^{\,\mathrm{cal,train}}$; this follows from the inequality and serves
only as a numerical check of the code.

\subsection{Component Dissection: One Significant Component}
\label{sec:results_dissection}

Table~\ref{tab:wilcoxon} reports all paired contrasts, and
Tables~\ref{tab:kendall_mae}, \ref{tab:acf_mae} and
\ref{tab:phys_viol_rate} the per-location means.  Apart from
\texttt{no\_calibration} (Sections~\ref{sec:results_floor}
and~\ref{sec:results_cal}), exactly one
component removal survives Holm correction on any metric:
\texttt{no\_D\_sorted} degrades Kendall $\tau$ MAE with a median paired
difference of $+\medNoDSortedTau$ ($p_{\mathrm{Holm}}\pHolmRelNoDSortedTau$).
The pre-declared pilot gate on Dubai, which required this effect, was
passed (Table~\ref{tab:pilot}).

\begin{figure}[t]
\centering
\includegraphics[width=\linewidth]{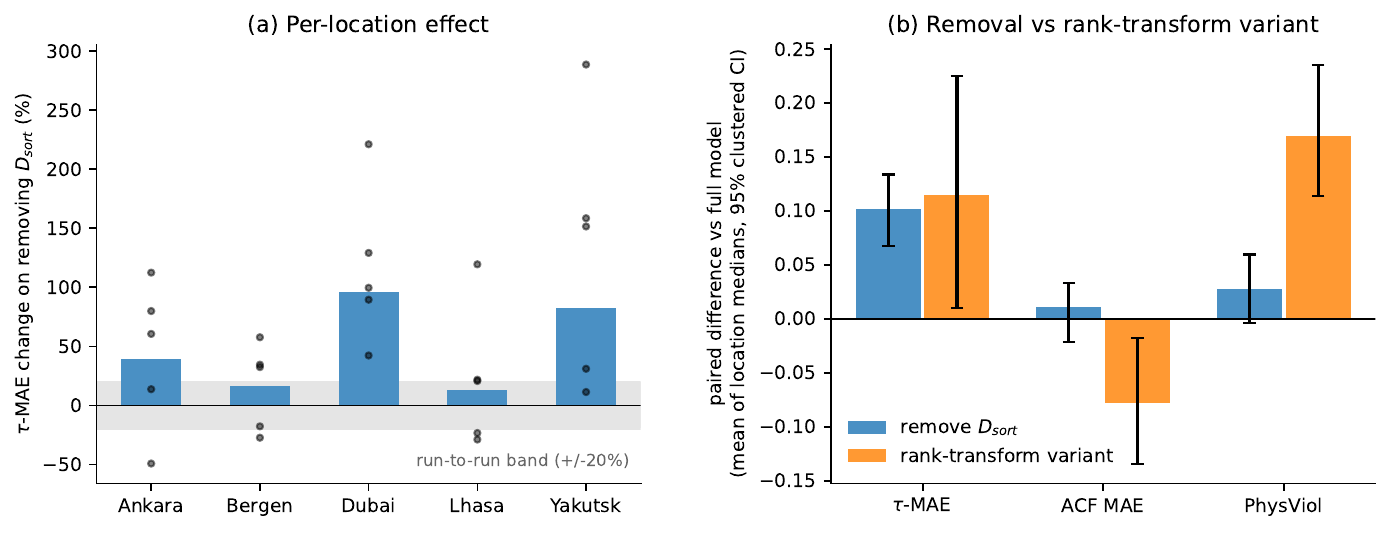}
\caption{The sorted-representation channel.  (a) Percentage change in
$\tau$-MAE on removing $\Dsort$, by location: bars are location means,
dots are seeds (each relative to the full model with the same seed), and
the grey band is the $\pm$\runNoiseBand\% run-to-run band.  The effect
is largest in Dubai and Yakutsk.  (b) Paired differences against the
full model for component removal (blue) and the rank-transformed
diagnostic variant (orange): bars are the mean over locations of the
per-location median difference, whiskers the \protoBootCI\%
location-clustered bootstrap interval (Table~\ref{tab:bootstrap}).  The
rank variant does not recover $\tau$ while improving ACF and raising
violations, consistent with quantile supervision.}
\label{fig:dsort}
\end{figure}

\paragraph{The effect varies in size across locations.}
Per-location mean degradation (Table~\ref{tab:kendall_mae} and
Fig.~\ref{fig:dsort}a) is \pctNoDSortedTauLhasa\% in Lhasa,
\pctNoDSortedTauBergen\% in Bergen, \pctNoDSortedTauAnkara\% in Ankara,
\pctNoDSortedTauYakutsk\% in Yakutsk and \pctNoDSortedTauDubai\% in
Dubai.  In Bergen and Lhasa the increase lies within the
$\pm$\runNoiseBand\% run-to-run band, with \nPosSeedsNoDSortedTauBergen{}
and \nPosSeedsNoDSortedTauLhasa{} of \nSeeds{} seeds positive; all seeds
degrade in Dubai and Yakutsk, and one Ankara seed improves
(Table~\ref{tab:wilcoxon}).  With \nLocsProtoWord{} locations we do not
attribute this variation to climate type; we report a component whose
effect has a consistent sign when pooled but a location-dependent size.

\paragraph{The rank-transformed variant is consistent with quantile supervision.}
The \texttt{rank\_D\_sort} arm is a diagnostic variant, not a component
of the model.  It replaces the column-wise sort with a
row-paired soft rank transform, so the discriminator observes the
empirical copula and the temporal order by construction while losing the
quantile values.  If
$\Dsort$ worked through copula information, this variant should retain
its benefit.  It does not (Fig.~\ref{fig:dsort}b): $\tau$ MAE degrades
by $+\medRankDSortTau$ ($p_{\mathrm{Holm}}\pHolmRelRankDSortTau$), comparable
in size to removing the component outright, although not in every
location (only one of five Bergen seeds is positive;
Table~\ref{tab:wilcoxon}).  The variant changes ACF
MAE by $\medRankDSortAcf$ ($p_{\mathrm{Holm}} = \pHolmRankDSortAcf$) and raises the physical
violation rate by $+\medRankDSortPhys$ ($p_{\mathrm{Holm}}\pHolmRankDSortPhys$), a
different profile: rank inputs keep temporal information
that sorted inputs discard, which may help autocorrelation, while the loss
of quantile-value supervision may hurt the marginals in physical units.
These results are consistent with the sorted representation acting as a
quantile-supervision channel for the generator during training, with its
effect on rank correlation metrics downstream of the marginal quantile
functions rather than of direct copula matching.  The arm remains in the
pre-specified Holm family of Table~\ref{tab:wilcoxon}, but it is read only
as this diagnostic (rule D4), not as a component (rule D2).

\subsection{Null Results}
\label{sec:results_ablation}

Every remaining contrast fails Holm correction on every metric
(Table~\ref{tab:wilcoxon}): removing $\Dtemp$
($\tau$ median $\medNoDTemporalTau$, $p_{\mathrm{Holm}} = \pHolmNoDTemporalTau$), removing $\Dstat$
($+\medNoDStatTau$, $p_{\mathrm{Holm}} = \pHolmNoDStatTau$), substituting the moment-vector
$\Dstat$ ($\medMomentDStatTau$, $p_{\mathrm{Holm}} = \pHolmMomentDStatTau$), removing the auxiliary
losses ($+\medNoAuxLossesTau$, $p_{\mathrm{Holm}} = \pHolmNoAuxLossesTau$), and fixing the critic
weights ($\medFixedLambdaTau$, $p_{\mathrm{Holm}} = \pHolmFixedLambdaTau$).  The
location-clustered bootstrap (Table~\ref{tab:bootstrap}), which is
descriptive only, does not contradict any Holm-significant contrast, and
no interval that excludes zero has the opposite sign to its Wilcoxon
median.  Its
intervals do, however, exclude zero for several contrasts that fail Holm
correction: \texttt{moment\_D\_stat} and \texttt{no\_aux\_losses} on
$\tau$-MAE, \texttt{no\_D\_stat} and \texttt{no\_calibration} on ACF MAE,
and \texttt{no\_D\_stat} and \texttt{no\_D\_temporal} on the violation
rate.  With \nLocs{} clusters and no multiplicity correction these
intervals do not override the pre-specified Wilcoxon and Holm rule.  At
\nSeeds{} seeds per cell and a
run-to-run noise of $\abs{\Delta\tau} = \runNoiseDtau$, effects of the
size these components could plausibly have are below the resolution of
the study; we report them as indistinguishable, not as proven
equivalent.

\subsection{Calibration Raises the Physics Violations; Projection
            Removes Them}
\label{sec:results_cal}

\begin{figure}[t]
\centering
\includegraphics[width=0.85\linewidth]{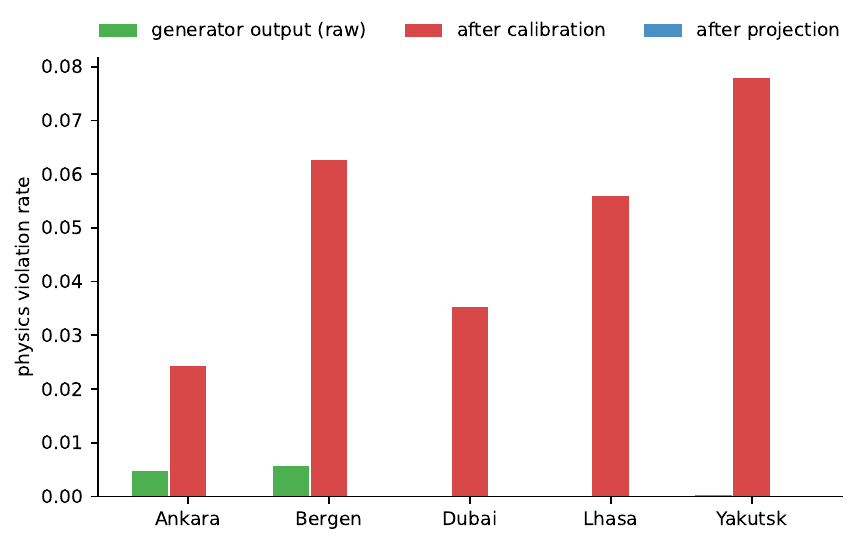}
\caption{Physical violations are raised by calibration and removed by
projection.  Full-model violation rates (mean over seeds) at three
pipeline stages: raw
generator output (green), after train-fitted isotonic calibration
(red), and after constraint projection (blue, exactly zero in all
runs).}
\label{fig:injection}
\end{figure}

The physical violation rate, measured before and after calibration,
localises the source of violations (Fig.~\ref{fig:injection} and
Table~\ref{tab:projection}).  Raw generator output is
nearly clean: full-model raw violation rates range from \physRawDubai{}
(Dubai and Lhasa) to \physRawBergen{} (Bergen).  After calibration they
range from \physCalAnkara{} (Ankara) to \physCalYakutsk{} (Yakutsk), and
the paired contrast confirms the direction
(\texttt{no\_calibration} minus full model $\medNoCalibrationPhys$,
$p_{\mathrm{Holm}}\pHolmNoCalibrationPhys$, with \texttt{no\_calibration}
cleaner in most pairs; Table~\ref{tab:wilcoxon}).  Each per-feature isotonic map is individually
monotone, but the maps are fitted independently, so jointly they can
move a (temperature, dewpoint) pair across the constraint boundary that
the generator respected.

Projecting calibrated output onto
$\mathcal{X}_{\mathrm{phys}}$ (Section~\ref{sec:calibration}) drives the
violation rate to exactly zero in all \nProjRuns{} full-model runs
(Table~\ref{tab:projection}).  Its distributional cost is small but not
uniformly within the pre-specified tolerance: the median change in KS is
\dKsMedianAll{} and the largest single-run change \dKsMaxLhasa{} (Lhasa),
and the largest change in a location exceeds the tolerance of
\protoDFiveFrac{} times the mean reference distance in
\phDKsLocsOverThrWord{} locations (\phDKsLocsOverThrList; post-freeze
descriptive).  Rule D5, which required zero violations together with a
change within this tolerance, therefore \verdictDFive.  The pipeline
includes projection as its final stage, and we do not describe its cost
as zero.

\subsection{Baselines Under the Held-Out Protocol}
\label{sec:results_baselines}

Table~\ref{tab:baselines} reports both re-implemented baselines under the
identical pipeline, tested against the full model as a separate Holm
family of \protoNBaselinesWord{} contrasts.  RCGAN serves as the
under-performing reference: it is worse than the full model on Kendall
$\tau$ MAE ($p_{\mathrm{Holm}}\pHolmBaseRcganTau$) and on the physical
violation rate ($p_{\mathrm{Holm}}\pHolmBaseRcganPhys$), and
indistinguishable on ACF MAE.  That this model is nonetheless
indistinguishable from StatD2GAN on calibrated KS
(Section~\ref{sec:results_floor}) is the
strongest single demonstration in the study that calibrated marginal metrics did not separate generators that dependence metrics separate.

TimeGAN is the informative comparison.  Under this re-implementation it
is significantly better than StatD2GAN on Kendall $\tau$ MAE
($p_{\mathrm{Holm}}\pHolmBaseTimeganTau$) and on ACF MAE
($p_{\mathrm{Holm}}\pHolmBaseTimeganAcf$), and indistinguishable on KS
and on the violation rate.  We therefore make
no claim that StatD2GAN is superior to either baseline.  Both baselines
were re-implemented for this study and trained with fixed budgets, so
these comparisons concern these implementations rather than the
published models.

\subsection{Sequence-Level Diagnostics}
\label{sec:results_seq}

\begin{figure}[t]
\centering
\includegraphics[width=\linewidth]{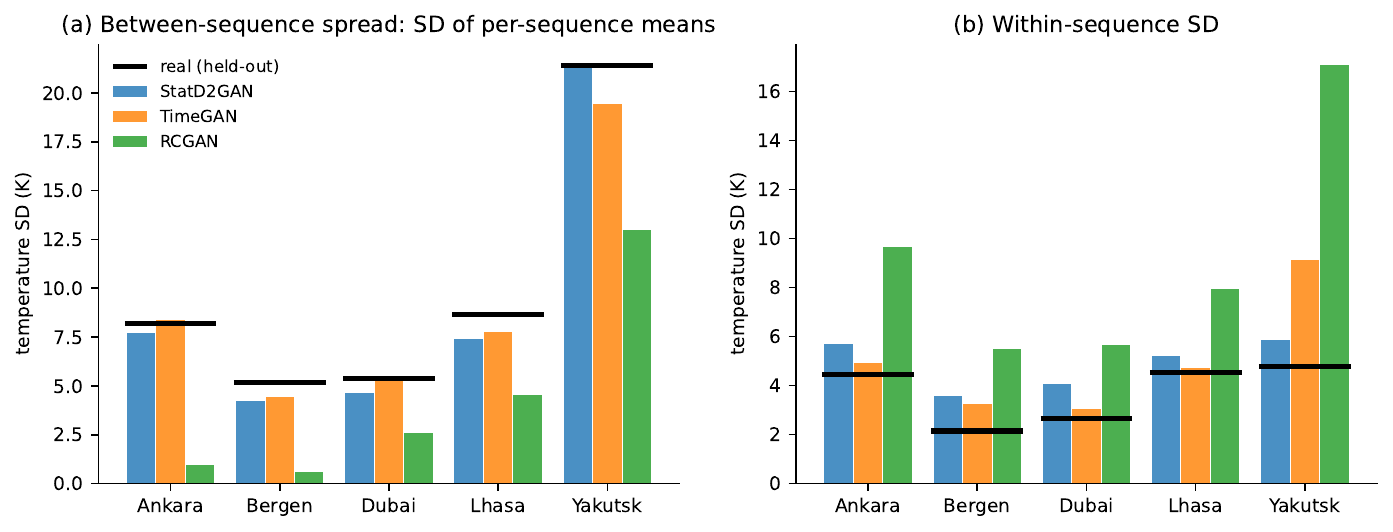}
\caption{Sequence-level temperature statistics.
(a)~Between-sequence spread: standard deviation of per-sequence weekly
temperature means.  (b)~Within-sequence standard deviation.  Bars: seed
medians for StatD2GAN (blue), TimeGAN (orange) and RCGAN (green); black
lines: real held-out values.}
\label{fig:variance}
\end{figure}

Table~\ref{tab:seq} and Fig.~\ref{fig:variance} decompose the
temperature variance that pooled metrics aggregate over.  The
between-sequence spread (Section~\ref{sec:metrics}: standard
deviation of per-sequence weekly means) is a proxy for climatic variability: sequences are not calendar-labelled, so it captures
between-sequence climatic variability rather than calendar seasonality
directly. In the real
held-out data, this spread ranges from \SeqSdRealBergen\,K (Bergen) to \SeqSdRealYakutsk\,K
(Yakutsk).
The pre-specified collapse criterion (rule D6: spread below
\protoDSixPct\% of the real value) was not met for either baseline at
the location level.  RCGAN comes closest, with \SeqSdRcganAnkara\,K in
Ankara and \SeqSdRcganBergen\,K in Bergen against \SeqSdRealAnkara\,K and
\SeqSdRealBergen\,K real, and it falls below the threshold in
\phSeedsBelowTenRcgan{} of \phSeedsPerModel{} individual runs (post-freeze
descriptive).  TimeGAN's spread is closer to the real one than
StatD2GAN's in Ankara, Bergen, Dubai and Lhasa, and further from it in
Yakutsk (\SeqSdTimeganYakutsk\,K against \SeqSdStatYakutsk\,K and
\SeqSdRealYakutsk\,K real).  StatD2GAN produces less between-sequence
spread than the real data in four locations (for example
\SeqSdStatBergen\,K against \SeqSdRealBergen\,K in Bergen) and nearly the
same in Yakutsk (\SeqSdStatYakutsk\,K against \SeqSdRealYakutsk\,K), and more
within-sequence standard deviation (\WithinSdStatBergen\,K against
\WithinSdRealBergen\,K in Bergen).  TimeGAN's within-sequence standard
deviation is closer to the real value than StatD2GAN's except in Yakutsk,
where it reaches \WithinSdTimeganYakutsk\,K against
\WithinSdRealYakutsk\,K real.

Pooled metrics do not expose this decomposition, because pooling over
sequences mixes between- and within-sequence variance.  RCGAN shows the
consequence: despite the compressed spread described above, its calibrated KS
and its ACF MAE are indistinguishable from StatD2GAN's
($p_{\mathrm{Holm}}\pHolmBaseRcganKs$ on KS; Table~\ref{tab:baselines}).
Kendall $\tau$ MAE ($p_{\mathrm{Holm}}\pHolmBaseRcganTau$) and the
violation rate ($p_{\mathrm{Holm}}\pHolmBaseRcganPhys$) separate it, and
only the sequence-level statistics show the compression.  Calibrated
marginal and pooled autocorrelation metrics are therefore not sufficient
on their own; we report dependence and sequence-level metrics alongside
them.

\FloatBarrier

\section{Discussion}
\label{sec:discussion}

\subsection{What the Sorted Representation Actually Does}

The dissection of Section~\ref{sec:results_dissection} replaces the
copula reading of Section~\ref{sec:discriminators} with a more modest
one that the evidence is consistent with.  The sorted representation hands the
discriminator each feature's within-sequence quantile function, and the
generator gradient that flows back through it supervises those quantile
functions directly, during training and in physical units, which
post-hoc calibration cannot do because it only remaps whatever the
generator produced.  Removing it degrades Kendall $\tau$ MAE in every
location on average, by \pctNoDSortedTauLhasa\% (Lhasa) to
\pctNoDSortedTauDubai\% (Dubai), with the Bergen and Lhasa increases
inside the $\pm$\runNoiseBand\% run-to-run band.  A quantile-supervision
channel should matter more where the unsupervised generator's quantiles
would land further from the truth, but with \nLocsProtoWord{} locations
this study cannot test that prediction, and we do not attribute the
location-dependent size of the effect to climate type.

\subsection{Calibration's Double Role, and What Evaluation Must Do
            About It}

Two findings jointly characterise isotonic calibration.  It saturates
the metric family it targets, bringing every calibrated configuration
close to the marginal noise-and-shift floor (full-model median
\ksRatioFullMedian{} times the floor), so that neither baseline,
including the under-performing RCGAN, is distinguishable from StatD2GAN
on it; and it raises physical constraint violation rates from at most
\physRawBergen{} in raw generator output to \physCalAnkara{} to
\physCalYakutsk{} after calibration.
By Proposition~\ref{prop:lock}, the locking to the floor is measured by
$\bar d^{\,\mathrm{cal,train}}$, and the architecture enters only through that
term.
Neither finding says calibration should be dropped: uncalibrated output
lies \ksRatioNoCalLhasa{} (Lhasa) to \ksRatioNoCalAnkara{} (Ankara) times
the floor, and the violations are removed by a projection whose
distributional cost is small but, contrary to rule D5, not uniformly
within the pre-specified tolerance (median change in KS \dKsMedianAll,
maximum \dKsMaxLhasa; Section~\ref{sec:results_cal}).  What the findings do say is that a
calibrated pipeline must not be evaluated on the statistics calibration
optimises, and must be checked for constraint damage after, not before,
the calibration stage.
Before calibration the marginals do differ across generators,
descriptively: the median raw KS over the \phSeedsPerModel{} runs of each
model is \phKsRatioRawTimeganMed{}, \phKsRatioRawFullMed{} and
\phKsRatioRawRcganMed{} times the floor for TimeGAN, StatD2GAN and RCGAN,
against \phKsRatioCalTimeganMed{}, \phKsRatioCalFullMed{} and
\phKsRatioCalRcganMed{} after calibration (post-freeze; no test is
attached), so before calibration TimeGAN fits the marginals more closely
than StatD2GAN, and calibration brings all three medians to within a
few per cent of the floor.

We suggest the following as minimum practice for calibrated generative
time series pipelines: fit all data-dependent stages, calibration
included, strictly on training data and evaluate on a temporally
embargoed held-out block; reference marginal metrics against a
real-versus-real noise-and-shift floor rather than a significance threshold;
attribute constraint violations to a pipeline stage by measuring before
and after each stage; and report matched-pair statistics with
multiplicity correction, with the run-to-run noise level stated so that
readers can judge what effect sizes the study can resolve.

\subsection{Reading the Null Results}

Of the \protoNContrastsWord{} contrasts, five (\texttt{no\_D\_temporal},
\texttt{no\_D\_stat}, \texttt{moment\_D\_stat}, \texttt{no\_aux\_losses}
and \texttt{fixed\_lambda}) are statistically indistinguishable from the
full model on every metric.  Read as architecture guidance, this is a parsimony
argument: of the three discriminators, the auxiliary losses, and the
evolutionary weighting, only the sorted-representation channel carries
demonstrable weight at this study's resolution, and a practitioner
constrained by compute could consider dropping $\Dstat$ and the replicator
machinery, bearing in mind that indistinguishability at \nSeeds{} seeds
per cell is not evidence of equivalence.  Read methodologically, it is a
caution about how multi-component architectures accumulate: each of
these components entered the design with a plausible rationale, and
none of the five produces an effect detectable under matched-pair
testing on a held-out split.

\subsection{Limitations}

Six limitations qualify the results.  First, with \nSeeds{} seeds per cell
(\nPairs{} matched pairs per contrast)
and run-to-run noise of $\abs{\Delta\tau} = \runNoiseDtau$, the study
resolves only large effects; the null results are statements of
indistinguishability at this resolution, not of equivalence.  Second,
all evidence comes from one data source, ERA5 reanalysis at
\nLocsProtoWord{} locations with \protoNScaledWord{} scaled
features; the protocol transfers to other domains, but the component
findings need not.  Third, the noise-and-shift floor is itself a function of the
train-to-validation distribution shift: Lhasa's floor (\ksFloorLhasa) is
the highest, against \ksFloorDubai{} to \ksFloorYakutsk{} elsewhere, so
floor-relative ratios compare against location-specific, not universal,
baselines.  Fourth, the constraint projection is boundary clipping; it
concentrates probability mass exactly on the constraint surface, which
is the source of the marginal change reported in
Section~\ref{sec:results_cal} and may matter more for downstream
uses sensitive to boundary atoms.  Fifth, the latent dimension (\protoLatent{}) and training length
(\protoEpochs{} epochs) were chosen by manual trials during development of
the generator, before the held-out protocol: smaller latent dimensions did
not train stably, and beyond \protoEpochs{} epochs (tested up to 400) the
training signal showed no further change. The remaining settings were set
during development by comparing parallel runs under an earlier evaluation
design, also before the held-out protocol. That design trained and
evaluated on the first 95\% of each record, which includes most of the
years later held out, so the held-out block is not independent of these
settings. None of the settings was re-tuned under the held-out protocol. All arms share
the same settings, which avoids differential tuning across ablations; however,
the settings were developed with the full model, so the removal contrasts
measure the effect of removing a component at settings that were not re-tuned
for the reduced model, and the size of component effects may depend on the
chosen hyperparameter regime. Sixth, RCGAN and TimeGAN are
compact re-implementations written for this study
(Section~\ref{sec:baselines}), trained with fixed budgets and not tuned,
so the comparisons concern these implementations rather than the
published models. StatD2GAN's settings were tuned during development
whereas the baselines were not, which if anything favours StatD2GAN.
The baselines are not a survey of current generators, and a
diffusion- or transformer-based generator was outside the scope of a
study whose claim is about evaluation methodology rather than
architectural superiority, and adding one under the same held-out,
matched-pair protocol is the natural next step.

\section{Conclusion}
\label{sec:conclusion}

We rebuilt the evaluation of a multi-discriminator GAN for synthetic
weather sequences around a held-out protocol with an embargo and
train-fitted calibration, because a circular design, with calibration
fitted on the evaluation data, cannot support architectural conclusions.
Under this protocol, with pre-specified decision rules, five
findings stand.  Isotonic calibration drives the marginal KS distance
to a median of \ksRatioFullMedian{} times a per-location real-versus-real
noise-and-shift floor, and neither re-implemented baseline, including the
under-performing RCGAN, is distinguishable from StatD2GAN on it, so calibrated
marginal metrics do not separate architectures at this resolution.  The
sorted-representation discriminator is the only significant
architectural component (Kendall $\tau$ MAE $+\medNoDSortedTau$ upon removal,
$p_{\mathrm{Holm}}\pHolmRelNoDSortedTau$), with a location-dependent effect from
\pctNoDSortedTauLhasa\% to \pctNoDSortedTauDubai\%, and a rank-transformed
diagnostic variant is consistent with its mechanism being quantile
supervision rather than copula matching.  Physical
constraint violations are raised by the calibration map rather than by the
generator, and constraint projection removes them with a marginal shift that
exceeds the pre-specified tolerance in \phDKsLocsOverThrWord{} locations.
The remaining components, the temporal and statistical discriminators,
the moment-based variant, the auxiliary losses and the evolutionary
weighting, are statistically
indistinguishable from the full model at this resolution, and we report those nulls
explicitly.  Finally, the calibrated marginal metric does not rank
generators that dependence metrics separate: the re-implemented TimeGAN
is significantly better than StatD2GAN on Kendall $\tau$ MAE and ACF MAE,
RCGAN is worse on Kendall $\tau$ MAE and constraint violations, and only
sequence-level statistics show RCGAN's compressed between-sequence
spread, although the pre-specified collapse criterion was not met for
either baseline.  StatD2GAN itself produces materially less between-sequence spread
than the real data in four of \nLocsProtoWord{} locations, and we make no
claim that it is superior to either baseline.

The architecture-specific findings may or may not transfer beyond ERA5.

\section*{CRediT authorship contribution statement}

\textbf{Mustafa Özaytaç:} Conceptualization, Methodology, Software,
Validation, Formal analysis, Investigation, Data curation, Writing --
original draft, Writing -- review \& editing, Visualization.
\textbf{\"{O}zge Karada\u{g} Ata\c{s}:} Supervision, Conceptualization,
Writing -- review \& editing.

\section*{Declaration of competing interest}

The authors declare that they have no known competing financial interests
or personal relationships that could have appeared to influence the
work reported in this paper.

\section*{Acknowledgements}

This study contains modified Copernicus Climate Change Service information
(2025). Neither the European Commission nor ECMWF is responsible for any use
that may be made of the information it contains.

\section*{Funding}

This research received no external funding.

\section*{Data availability}

The code (StatD2GAN implementation, baseline re-implementations, the
frozen evaluation protocol, analysis and figure-generation scripts) and
the result files (per-run metrics, statistical test outputs,
decision-rule evaluations, floor references) supporting the
findings of this study are openly available.
Quantities marked post-freeze in the text, Figs.~\ref{fig:location_robustness},
\ref{fig:dsort} and~\ref{fig:variance}, the reprints of Tables~\ref{tab:seq} and~\ref{tab:pilot}, and
Table~\ref{tab:dist} are
produced by \texttt{report\_extra.py}, \texttt{fig1\_extra.py}, \texttt{figs\_extra.py},
\texttt{table\_seq\_extra.py}, \texttt{table\_pilot\_extra.py} and \texttt{distance\_extra.py}, which
are released separately from the frozen evaluation code.
\texttt{distance\_extra.py} re-runs inference from the stored generator
weights with the frozen seeds and frozen code, without training
(\ref{app:distance}). The campaign was trained with torch
\phTrainTorch{} (run environment record); the recomputation used torch \phDistTorch{} on
the same GPU model, with a maximum difference of \phDistGateMaxDiff{} to the
stored KS, and bitwise agreement on other devices is not claimed;
Table~\ref{tab:dist} is computed from its output.
\texttt{report\_extra.py} additionally reads the output of
\texttt{distance\_extra.py}; the remaining scripts read only frozen outputs. They are outside the frozen
protocol and carry no test.  The frozen and post-freeze layers are archived together on Zenodo
at \url{https://doi.org/10.5281/zenodo.22996771}, with a GitHub
mirror at \url{https://github.com/ozaytac/StatD2GAN}.
The 275 trained generator weight files read by \texttt{distance\_extra.py} are archived in a
companion Zenodo record, \url{https://doi.org/10.5281/zenodo.23480462}.



\appendix
\section{Origin of the Default Hyperparameters}
\label{app:sensitivity}

Table~\ref{tab:hyperparams}'s latent dimension (\protoLatent{}) and training
length (\protoEpochs{} epochs) were chosen by manual trials during development
of the generator, before the held-out protocol, on the basis of training
behaviour: smaller latent dimensions did not train stably, and beyond
\protoEpochs{} epochs (tested up to 400) the training signal showed no further
change. The remaining settings were compared on parallel runs under an
earlier evaluation design in which the generator was trained on the first
95\% of each record and calibration and metrics were computed against that
same training data. Because that design is exactly the circularity this
paper's protocol removes, we do not present it as a sensitivity result: no
figure, table or numeric claim depends on it, and it is stated here only as
provenance. A reader applying the method to other data should re-derive
these settings under the held-out protocol.

\section{Distance of the Calibrated Sample to the Training Block}
\label{app:distance}

Proposition~\ref{prop:lock} involves $\bar d^{\,\mathrm{cal,train}}$, which
the frozen run records do not contain.  The post-freeze script
\texttt{distance\_extra.py} reloads the stored generator weights of the
\phDistNRuns{} calibrated runs (\texttt{no\_calibration} excluded), redraws the
evaluation and calibration samples with the frozen seeds, refits the
calibration map with the frozen code and computes the per-feature distances.
The distances were recomputed on the GPU model recorded in the run records
(\phDistDevice{}, torch \phDistTorch{}); the maximum difference to the stored
KS was \phDistGateMaxDiff{} over \phDistNRuns{} runs (tolerance
$\phDistGateTol$).  Table~\ref{tab:dist} is descriptive and carries no test.
The three largest ratios $\bar d^{\,\mathrm{cal,train}}/\mathrm{KS}^{\mathrm{floor}}$
are \phDistTopOneRatio{} (\phDistTopOneLoc{}, \phDistTopOneModel{}, seed
\phDistTopOneSeed{}), \phDistTopTwoRatio{} (\phDistTopTwoLoc{},
\phDistTopTwoModel{}, seed \phDistTopTwoSeed{}) and \phDistTopThreeRatio{}
(\phDistTopThreeLoc{}, \phDistTopThreeModel{}, seed \phDistTopThreeSeed{}).

\begin{table}[!htb]
\centering\small
\caption{Distance of the calibrated synthetic sample to the training block (post-freeze; descriptive, no tests; $\bar d^{\,\mathrm{raw,train}}$ is not a quality ranking). Medians over runs unless marked max. Location rows pool the calibrated ablation arms; baseline rows pool locations. $\bar d^{\,\mathrm{cal,train}}$, $\bar d^{\,\mathrm{raw,train}}$: feature-mean sup-distance of the calibrated and uncalibrated sample to the training block; floor: $\mathrm{KS}^{\mathrm{floor}}$; $\Delta=\lvert\mathrm{KS}^{\mathrm{cal}}-\mathrm{KS}^{\mathrm{floor}}\rvert$.}
\label{tab:dist}
\begin{tabular}{lrrrrrrr}
\toprule
 & Runs & Floor & $\bar d^{\,\mathrm{cal,train}}$ & \multicolumn{2}{c}{$\bar d^{\,\mathrm{cal,train}}/$floor} & $\Delta/$floor & $\bar d^{\,\mathrm{raw,train}}$ \\
\cmidrule(lr){5-6}
 & & & & median & max & & \\
\midrule
Ankara & 40 & 0.0363 & 0.0056 & 0.154 & 0.422 & 0.032 & 0.200 \\
Bergen & 40 & 0.0360 & 0.0053 & 0.147 & 0.475 & 0.025 & 0.185 \\
Dubai & 40 & 0.0344 & 0.0067 & 0.194 & 0.512 & 0.029 & 0.131 \\
Lhasa & 40 & 0.0846 & 0.0059 & 0.069 & 0.223 & 0.010 & 0.165 \\
Yakutsk & 40 & 0.0488 & 0.0075 & 0.153 & 0.346 & 0.043 & 0.179 \\
\midrule
RCGAN & 25 & -- & 0.0023 & 0.043 & 0.428 & 0.008 & 0.437 \\
TimeGAN & 25 & -- & 0.0064 & 0.137 & 0.699 & 0.035 & 0.079 \\
\midrule
All runs & 250 & -- & 0.0059 & 0.134 & 0.699 & 0.023 & 0.172 \\
\bottomrule
\end{tabular}
\end{table}

\FloatBarrier

\section*{Declaration of generative AI and AI-assisted technologies in
the manuscript preparation process}

During the preparation of this work the authors used Claude (Anthropic) to
assist with manuscript drafting and revision, LaTeX formatting and structural
edits, consistency checks of statistical reporting across the text, tables and
figures, and the writing of the analysis, reporting and figure scripts and
verification tools released with the paper. After using this tool, the authors
reviewed and edited the content as needed and take full responsibility for the
content of the published article.

\bibliographystyle{elsarticle-num}

\end{document}